\documentclass{article}

\usepackage{microtype}
\usepackage{graphicx}
\usepackage{subcaption}
\usepackage{booktabs} %

\usepackage{hyperref}
\usepackage{graphicx}
\usepackage{amsmath}
\usepackage{amssymb}
\usepackage{booktabs}

\usepackage{amsmath,amsfonts,bm}

\def\eqref#1{equation~\ref{#1}}

\def\1{\bm{1}}

\DeclareMathAlphabet{\mathsfit}{\encodingdefault}{\sfdefault}{m}{sl}
\SetMathAlphabet{\mathsfit}{bold}{\encodingdefault}{\sfdefault}{bx}{n}

\usepackage{url}

\usepackage{graphicx, color}
\usepackage{tikz}
\usepackage{comment}
\usepackage{amsmath,amssymb} %
\usepackage{color}
\usepackage{ragged2e}
\usepackage{xcolor}         %
\usepackage{graphicx}
\usepackage{multirow}

\usepackage{subcaption}

\usepackage{caption}
\usepackage{float}

\usepackage{url}            %
\usepackage{booktabs}       %
\usepackage{amsmath}
\usepackage{amsfonts}       %

\usepackage{nicefrac}       %
\usepackage{microtype}      %
\usepackage{nicefrac}       %
\usepackage{wrapfig}

\usepackage{multirow}
\usepackage{makecell}
\usepackage{threeparttable}
\usepackage{sidecap}
\usepackage{caption}

\newcommand{\btheta}{\boldsymbol{\theta}}
\newcommand{\bphi}{\boldsymbol{\phi}}

\usepackage{colortbl} %
\definecolor{mColor1}{rgb}{0.9,0.9,0.9}
\definecolor{mColor2}{rgb}{0.95,0.95,0.95}
\definecolor{non-photoblue}{rgb}{0.64, 0.87, 0.93}
\definecolor{lightblue}{rgb}{0.81, 0.94, 1.0}
\usepackage{tikz}
\usepackage{pifont}
\newcommand{\ymark}{\ding{51}}

\definecolor{mColor1}{rgb}{0.9,0.9,0.9}
\definecolor{mColor2}{rgb}{0.95,0.95,0.95}
\definecolor{non-photoblue}{rgb}{0.64, 0.87, 0.93}
\definecolor{lightblue}{rgb}{0.81, 0.94, 1.0}
\definecolor{lightorange}{rgb}{0.965, 0.835, 0.71}

\definecolor{mygreen}{rgb}{0.000, 0.392, 0.000}

\usepackage{orcidlink}
\definecolor{mypurple}{rgb}{0.502, 0.000, 0.502}

\newcommand{\lf}[1]{\textcolor{black}{#1}}
\newcommand{\lfs}[1]{\textcolor{black}{#1}}
\newcommand{\lff}[1]{\textcolor{black}{#1}} 
\newcommand{\lflf}[1]{\textcolor{black}{#1}}

\usepackage{algorithm}
\usepackage{algorithmic}
\usepackage{pifont}

\usepackage[capitalize]{cleveref}
\crefname{section}{Sec.}{Secs.}
\Crefname{section}{Section}{Sections}
\Crefname{table}{Table}{Tables}
\crefname{table}{Tab.}{Tabs.}

\usepackage{booktabs} %
\usepackage{multirow} %
\usepackage{graphicx} %

\usepackage{soul}
\usepackage{xcolor}

\sethlcolor{yellow} %

\definecolor{cgreen}{rgb}{0.2,0.6,1}

\definecolor{darkgreen}{RGB}{0,100,0}

\usepackage{etoc}

\usepackage{amsmath,amssymb}
\usepackage{bm} %
\newcommand{\bx}{\mathbf{x}}
\newcommand{\by}{\mathbf{y}}
\newcommand{\bz}{\mathbf{z}}
\newcommand{\bp}{\mathbf{p}}

\newcommand{\balpha}{\bm{\alpha}}

\usepackage{colortbl} %
\definecolor{mColor1}{rgb}{0.9,0.9,0.9}
\definecolor{mColor2}{rgb}{0.95,0.95,0.95}
\definecolor{non-photoblue}{rgb}{0.64, 0.87, 0.93}
\definecolor{lightblue}{rgb}{0.81, 0.94, 1.0}
\definecolor{lightorange}{rgb}{0.965, 0.835, 0.71}

\usepackage[preprint]{icml2026}

\usepackage{amsmath}
\usepackage{amssymb}
\usepackage{mathtools}
\usepackage{amsthm}
\usepackage{enumitem}

\theoremstyle{plain}

\theoremstyle{definition}

\theoremstyle{remark}

\usepackage[textsize=tiny]{todonotes}

\icmltitlerunning{The Mean is the Mirage: Entropy-Adaptive Model Merging under Heterogeneous Domain Shifts in Medical Imaging}

\begin{document}

\twocolumn[
  \icmltitle{The Mean is the Mirage: Entropy-Adaptive Model Merging \\under Heterogeneous Domain Shifts in Medical Imaging}

  \icmlsetsymbol{equal}{*}

        \begin{icmlauthorlist}
        
        \icmlauthor{Sameer Ambekar}{tum,helm,relai,mcml}
        \icmlauthor{Reza Nasirigerdeh}{path,mcml}
        \icmlauthor{Peter J. Sch\"uffler}{path,tum,mcml}
        \icmlauthor{Lina Felsner}{tum}
        \icmlauthor{Daniel M. Lang}{tum,helm,equal}
        \icmlauthor{Julia A. Schnabel}{tum,helm,relai,mcml,kcl,equal}
        
        \end{icmlauthorlist}
        
        \icmlaffiliation{tum}{School of Computation, Information and Technology, Technical University of Munich, Germany}
        \icmlaffiliation{helm}{Institute of Machine Learning in Biomedical Imaging, Helmholtz Munich, Germany}
        \icmlaffiliation{relai}{relAI – Konrad Zuse School of Excellence in Reliable AI}
        \icmlaffiliation{mcml}{Munich Center for Machine Learning (MCML)}
        \icmlaffiliation{kcl}{School of Biomedical Engineering and Imaging Sciences, King's College London, UK}
        \icmlaffiliation{path}{Institute of Pathology, Technical University of Munich, Germany}
        
        \icmlcorrespondingauthor{Sameer}{firstandlastnameatttum.de}
  \vskip 0.3in
]
\printAffiliationsAndNotice{\icmlEqualContribution}

\begin{abstract}

Model merging under unseen test-time distribution shifts often renders naive strategies, such as mean averaging unreliable. This challenge is especially acute in medical imaging, where models are fine-tuned locally at clinics on private data, producing domain-specific models that differ by scanner, protocol, and population. When deployed at an unseen clinical site, test cases arrive in unlabeled, non-i.i.d. batches, and the model must adapt immediately without labels. In this work, we introduce an entropy-adaptive, fully online model-merging method that yields a batch-specific merged model via only forward passes, effectively leveraging target information. We further demonstrate why mean merging is prone to failure and misaligned under heterogeneous domain shifts. Next, we mitigate encoder classifier mismatch by decoupling the encoder and classification head, merging with separate merging coefficients. We extensively evaluate our method with state-of-the-art baselines using two backbones across nine medical and natural-domain generalization image classification datasets, showing consistent gains across standard evaluation and challenging scenarios. These performance gains are achieved while retaining single-model inference at test-time, thereby demonstrating the effectiveness of our method.

\end{abstract}

\section{Introduction}
Recent advances in transfer learning using pretrained models have driven deep learning to achieve strong performance with limited training data~\cite{he2016deep,lora_iclr2022,gpt3_2020}.
This is particularly valuable in medical imaging, where the model training on large datasets is constrained by the high cost of expert annotation and strict privacy regulations~\cite{hoofnagle2019european,kaissis2020secure}.
Therefore, each clinical site typically fine-tunes a pretrained encoder locally, 
resulting in multiple {clinical site} domain-specific models which are nearly oracle models on their data. These clinical sites do not share the data with one another due to privacy concerns~\cite{zhu2024privacy} and to capture features specific to their scanners, acquisition protocols, and patient populations.
\begin{figure}[t]
\centering
\includegraphics[width=1.0\linewidth, height=1.0\textheight, keepaspectratio]{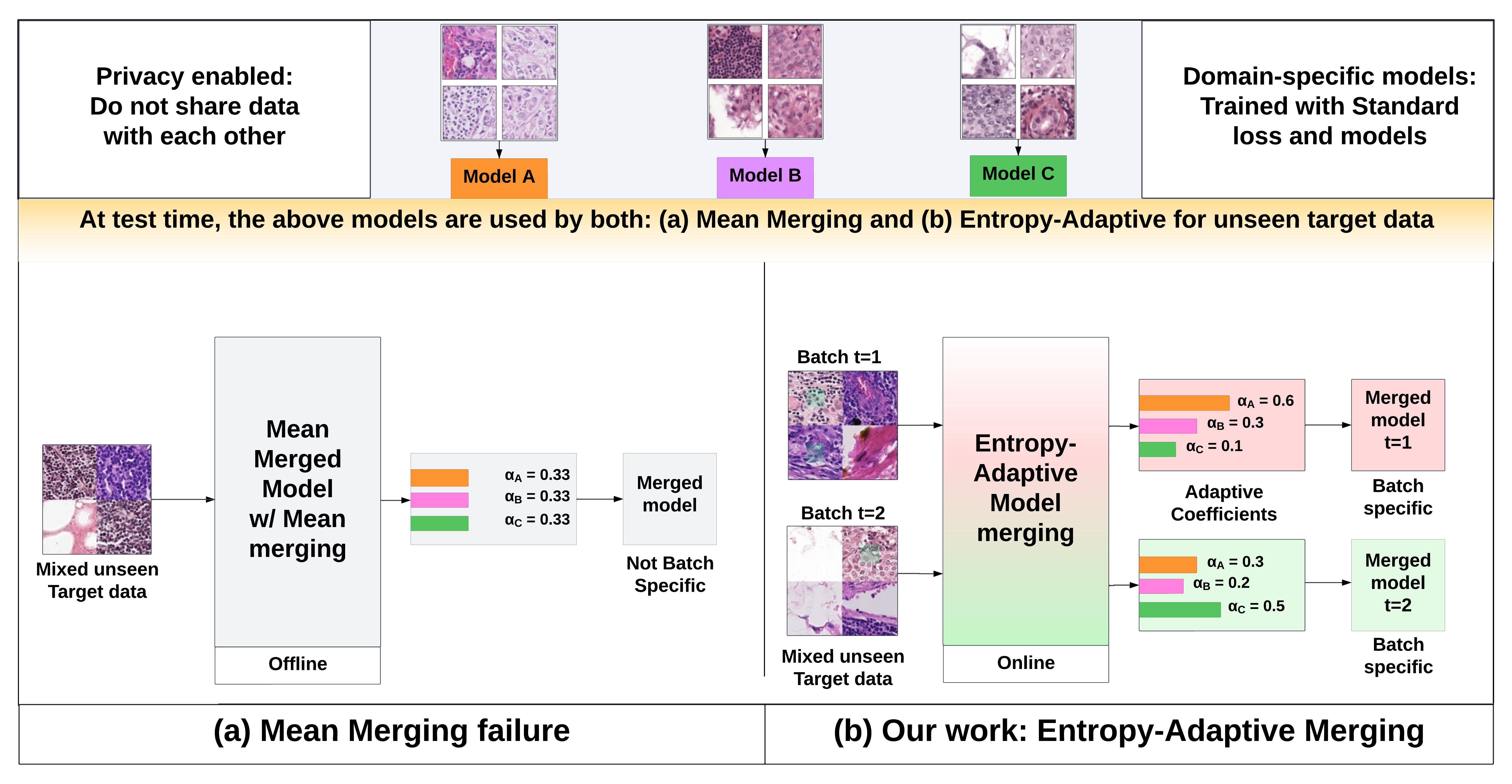}

\caption{\textbf{Mean merging and Entropy-Adaptive merging (Ours).} \textbf{(a)} Mean merging averages independently trained hospital models into a static model that doesn't consider the target information into account, thus can fail on mixed target batches.
\textbf{(b)} Our method, adaptively, uses entropy for each unlabeled target batch to calculate per-batch merging coefficients along linear mode connectivity directions, producing batch-specific merged models while leveraging target information. Additional illustration with loss heatmaps in Fig.\ref{fig:weight_spaces}. 
}

\label{fig1:all}
\vspace{-4mm}
\end{figure}

A \lfs{central} question is whether these independently trained domain-specific models (cross-clinic models) can be reused during deployment at an unseen hospital. This is especially interesting and important in scenarios where this target-domain deployment is inherently {online}, \lfs{i.e.} diverse unlabeled cases arrive incrementally in batches, and the system must respond immediately as the stream evolves, a setting reminiscent of test-time adaptation~\citep{wang2021tent}. 
With multiple expert models, the practical difficulty is that adapting all \lfs{models} for each batch at test time is operationally undesirable with backpropagation, due to the high computational cost of updating multiple domain-specific models~\cite{liang2020we}.
Instead, %
a single model that can be {selected or personalized per incoming batch} using only unlabeled target data is desirable.

A naive strategy is to select a single domain-specific model from the multiple domain-specific models for every batch, but this is prone to failure when the target distribution is unknown, changes over time, or contains mixed cases~\cite{wang2022continual,niu2023towards,zhao2023pitfalls}.
Model ensembling can improve robustness, but requires multiple forward passes (higher memory and compute) and can dilute specialization: for a given sample, one expert may be clearly most appropriate, but uniform averaging of predictions mixes it with less relevant experts~\cite{SnapshotEnsembles,sagi2018ensemblesurvey}.
Recently, model merging~\cite{Modelsoups_ICML2022,FisherMerging_NeurIPS2022,ReBasin_ICLR2023,georgestoicaZipItMergingModels2023,TiesMergingNeurIPS2023}
 offers a more deployable alternative by consolidating multiple expert models into a single checkpoint that runs with a single forward pass.
Mean averaging~\cite{frankle2020linear, Modelsoups_ICML2022} as in Fig.~\ref{fig1:all}a is the most common approach to merge multiple models, which is training-free and can work when models are highly correlated and closely aligned~\cite{Modelsoups_ICML2022,modelfusion_survey2023}. These methods compute an average across all available model weights to obtain a single merged model for the target batch. 

However, \lfs{this naive approach is problematic in inherently complex real-world settings, such as medical imaging.}
Even if the domain-specific models share the same initialization, they are fine-tuned independently on heterogeneous, non-overlapping cross-domain data.
Consequently, at test time, only a subset of layers may remain aligned or similar~\cite{lee2022surgical}, while others may drift in domain-specific ways (e.g., due to scanner/protocol appearance shifts).
This \lfs{makes mean} merging only partially valid: it may work for some layers, but applying it uniformly across all layers can produce a functionally inconsistent network even when each expert performs well on its own hospital. (See Fig.~\ref{fig1:all}a)

Recently, this observation has motivated {adaptive} merging strategies that go beyond \lfs{mean} averaging, including task-vector merging as in TIES-Merging~\cite{TiesMergingNeurIPS2023}, permutation based merging~\cite{ReBasin_ICLR2023}, and approaches that learn layer-wise or task-wise weighting coefficients~\cite{daheim2023model,emrmerging_arxiv2024,tangFusionBenchComprehensiveBenchmark2024,khan2024sok,yadavSurveyModelMoErging2024}. However, the key limitations for medical deployment scenarios are that they are either effective {offline} (they produce one merged model throughout the target domain) or rely on ad-hoc backpropagation on target data with multiple hyperparameters~\cite{niu2024test,AdaMerging_ICLR2024}.

\lfs{In this work, w}e propose a {fully online}, {batch-specific,} 
\lfs{{Entropy-Adaptive linear connectivity merging} mechanism,}
that selects merging coefficients along linear connectivity directions as shown in Fig.~\ref{fig1:all}b.
The main advantage of our method is that it selects a merged model from feed-forward signals alone, without updating domain-specific model weights or using test-time backpropagation, making it suitable for real-time clinical workflows with limited on-site compute and strict latency constraints.
Our method is designed for the fully online setting: unlabeled cases arrive sequentially, and importantly, each incoming batch can itself be heterogeneous (e.g., mixed protocols, patient subpopulations, or acquisition conditions). %
We therefore compute the merging coefficients that define a \emph{personalized} merged model for the current batch, and repeat this forward-only selection as the stream evolves. We use entropy as an unsupervised signal, \lfs{where} lower entropy on the current batch serves as an unsupervised signal for merging.
This is particularly advantageous when the current batch is closer to one expert \lfs{model} than the others: entropy naturally emphasizes the expert that yields more confident predictions on that batch and down-weights less relevant domain-specific models, while allowing this preference to change from batch to batch as the target stream evolves. Importantly, the same approach can blend layers that remain compatible due to shared pretraining while avoiding harmful averaging across layers that drift across cross-domain hospital data, preventing the uncertainty collapse often observed with mean merging. 
{In summary, we make the following three contributions:}
\begin{itemize}

\item {Entropy-Adaptive online merging:} We introduce an entropy-adaptive merging method that obtains merge coefficients from unlabeled target batches adaptively. This yields a batch-specific merged model that blends compatible layers while down-weighting incompatible domain-specific models adaptively at test time.
\item {Mean-merging failure analysis:} We show that mean merging is only partially valid, beneficial for layers that remain aligned, yet prone to sharp failures when applied uniformly across all layers. Additionally, we investigate this with systematic failure cases and layer-wise analyses.
\item {{Decoupled encoder and classification head coefficients for robust online merging:} 
We first demonstrate that fixed-mean merging can fail sharply due to misalignment between the classification head and the encoder under heterogeneous shifts. To address this failure mode, we obtain {module-decoupled} merge coefficients efficiently for encoder and head weights to avoid head-induced merging failures. }
\end{itemize}

We extensively validate our proposal across both medical and natural vision domain shift benchmarks, spanning 9 datasets and 2 ViT backbones, through comprehensive experiments.

\section{Related work }

In this section, we review a range of model-merging techniques and explain how the demands of our setting motivate a fundamentally different merging objective. 

\noindent \textbf{Mean-based model averaging.}
Model merging has emerged as a powerful paradigm for combining knowledge from multiple independently (pre-) trained models without requiring access to source data or extensive retraining~\cite{modelfusion_survey2023,yadavSurveyModelMoErging2024,khan2024sok,yang2024model,tangFusionBenchComprehensiveBenchmark2024}.
The simplest approach is mean averaging of trained models, which has been extensively studied in both theoretical and practical contexts~\cite{izmailovAveragingWeightsLeads2019,Modelsoups_ICML2022,kaddour2022flat}.
\citet{Modelsoups_ICML2022} introduced model soups, demonstrating that averaging weights of multiple fine-tuned models of the same architecture improves accuracy without increasing inference cost.
This idea has been extended to specialized scenarios, including sparse models~\cite{SparseModelSoups}, adapter modules~\cite{adaptersoup_EACL2023}, and diverse pre-training objectives~\cite{rame2023modelratatouille}.
However, naive averaging during inference often leads to parameter interference when models are trained on heterogeneous data~\cite{TiesMergingNeurIPS2023,yadav2024matters}.

\noindent \lfs{\textbf{Adaptive merging.}} %
\lfs{Recently, \citet{TiesMergingNeurIPS2023} proposed to resolve parameter interference with TIES-Merging, building on trimming redundant parameters, electing signs, and merging only parameters with aligned signs.}
Recent methods tackle heterogeneity through Fisher-weighted averaging~\cite{FisherMerging_NeurIPS2022}, gradient matching~\cite{daheim2023model,daheim2024model}, optimal transport~\cite{imfeld2024transformer}, and magnitude-based sampling~\cite{deep2024della}.
For multi-task scenarios, several adaptive methods have been proposed.
AdaMerging~\cite{AdaMerging_ICLR2024} with backpropagation adaptively determines layer-wise merging coefficients using test data, while~\citet{tangMergingMultiTaskModels2024,shen2024efficient} propose weight-ensembling mixture of experts for multi-task merging.
\citet{yangRepresentationSurgeryMultiTask2024} introduce representation surgery to eliminate task interference, and~\citet{tang2023concrete} use concrete subspace learning for multi-task model fusion.

\noindent\textbf{Model merging for different downstream tasks.} %
For medical imaging specifically, MedMerge~\cite{almakky2024medmerge} explores merging during supervised fine-tuning with labels from the target domain.
Other specialized applications include vision transformers~\cite{ye2023merging}, diffusion models~\cite{DiffusionSoup2024}, LLM alignment~\cite{tekin2024h,chegini2024model}, and cross-task generalization~\cite{ZipIt_Arxiv2023,georgestoicaZipItMergingModels2023,kim2025testtime}.

\noindent {Different from these model merging works, } we consider a multi-domain classification  setting in which domain-specific models are trained on disjoint domains and are misaligned with one another. To address this, we obtain target-specific merged weights online, producing a single merged model per batch without accessing labels or performing any gradient-based backpropagation at test time.

\begin{figure*}[t]
\centering
\includegraphics[width=1.0\linewidth, height=0.8\textheight, keepaspectratio]{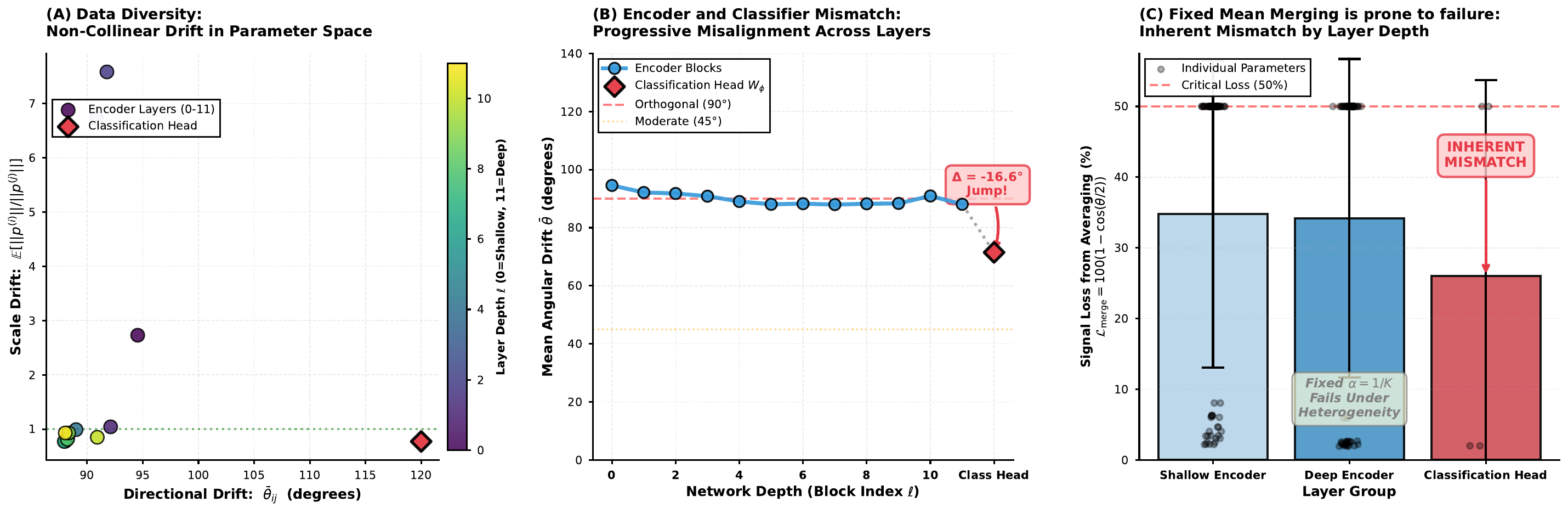}

\caption{{Why Mean Merging fails} for the PACS dataset using ViT-B32 domain-specific models trained on Photo, Cartoon, and Sketch domains and evaluated on Art domain.
We first compute the angles as provided in~\cite{jang2024model} between weights and bias terms of domain-specific trained models, then compute directional and angular drift to obtain deeper insights.
\textit{{(A) Data diversity:}} Model parameters drift in different directions and scales across domains, with the classification head showing the strongest misalignment. 
\textit{{(B) Encoder-classification head mismatch:}} \lflf{Mean angular layerwise drift.} Misalignment increases with network depth and becomes most pronounced at the classification head. 
\textit{{(C) Fixed mean merging is prone to failure:} } Mean merging leads to substantial signal loss in the classification head compared with individual models. 
}
\label{fig2:heterogeneity_analysis}
\end{figure*}

\section{Background and Notations}

We consider a single-domain training setup in which each source model is trained on labeled data from exactly one domain, without data sharing between models. During source training, we denote the $k$-th source domain as $D_s^{(k)}$ for $k \in \{1,\ldots,K\}$, with $K$ the number of source domains. Each domain's data obtained from medical centers can differ due to scanner manufacturer, acquisition protocol, and geographic or laboratory factors (e.g., staining and tissue preparation in pathology)~\cite{li2020domain,sun2022camelyon}.
The domain $k$ provides labeled samples $(\bx_s^{(k)},\by_s^{(k)})\sim\mathcal{D}_s^{(k)}$, and trains models {only} on its local data (no cross-hospital data sharing).
Each domain-specific model is parameterized as an encoder ($g(\bx;\btheta)$) followed by a linear classification head ($\phi(\bz)$):

\lf{
\begin{equation}
\begin{aligned} 
\label{eq:expert_form}
f_k(\bx;\btheta_s^{(k)}, W_{\bphi}^{(k)})
&= \bphi^{(k)}(g(\bx;\btheta_s^{(k)}))\\
&= (W_{\bphi}^{(k)})^\top g(\bx;\btheta_s^{(k)}) \lf{\,\,},
\end{aligned}
\end{equation}
}
where $\btheta_s^{(k)}$ are encoder weights and $W_{\bphi}^{(k)}\in\mathbb{R}^{d\times C}$ are classification head weights for $C$ classes. 

At test time or inference, we encounter unseen target hospital data from an unseen domain $D_t$ and observe unlabeled target batches $\{\mathcal{B}_t\}_{n=1}^N$ online at time point $n$, with $\bx\sim\mathcal{D}_t$. 
Here, the task is to use domain-specific models for each incoming target batch and obtain an ideal {single} deployed model at each time step, with minimal computation over target batch $\mathcal{B}_t$. To train the source models, we follow the Diverse Weight Averaging (DiWA)  setup~\cite{rame2022diverse} and perform multiple training runs for each domain, varying hyperparameters and procedures to promote functional diversity among models.

\section{Methodology}

In this section, we show why mean merging \lflf{is prone to failure and misalignment} under domain heterogeneity (Sec.~\ref {section:mean_brittle}) and then present our entropy-adaptive merging for adaptive online inference (Sec.~\ref {section:entropy_merging}).

\subsection{Why mean merging is prone to failure and misalignment}
\label{section:mean_brittle}
\begin{figure}[t]
\centering

\begin{subfigure}[t]{0.49\linewidth}
\centering
\includegraphics[width=\linewidth]{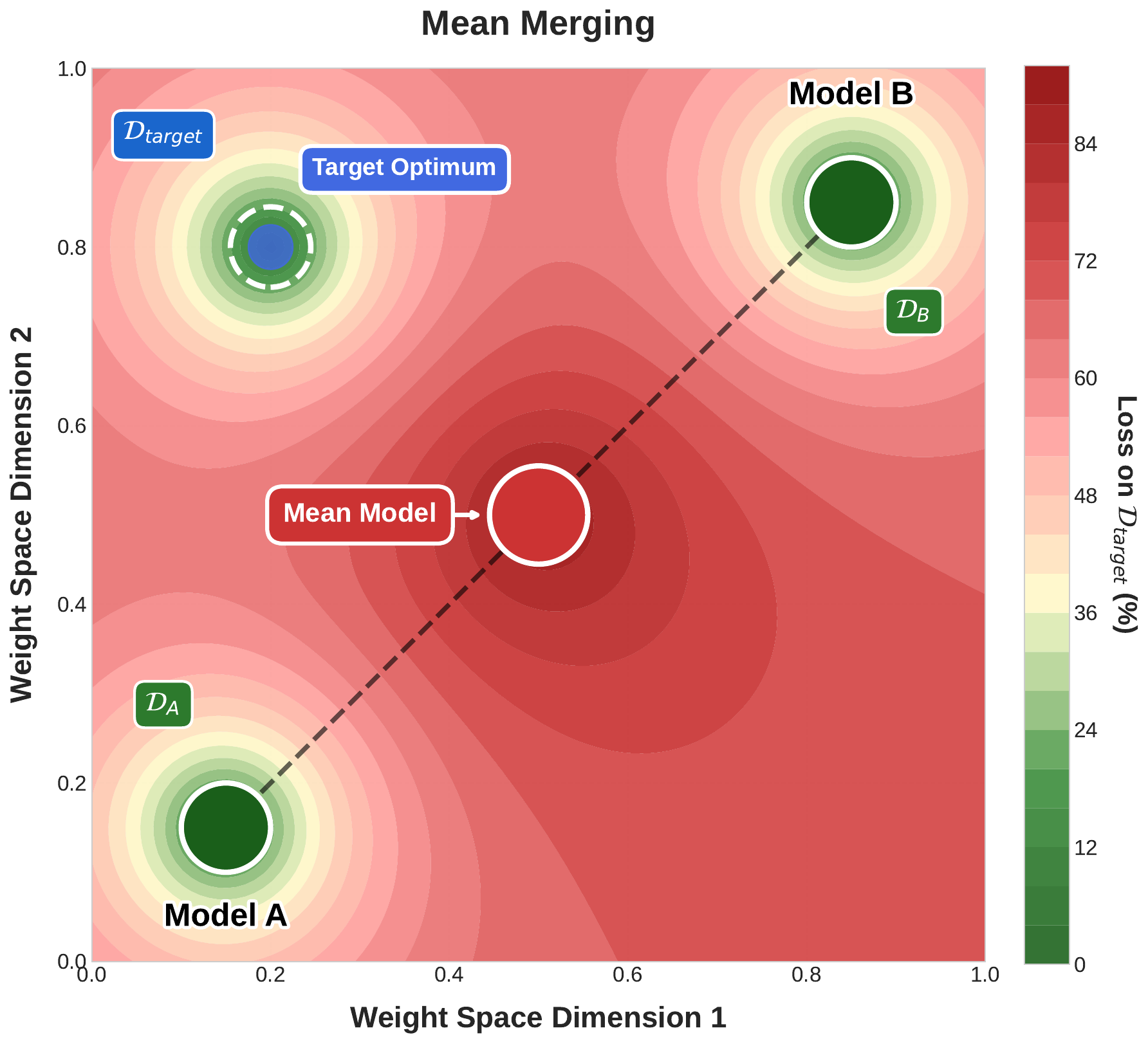}
\end{subfigure}
\hfill
\begin{subfigure}[t]{0.49\linewidth}
\centering
\includegraphics[width=\linewidth]{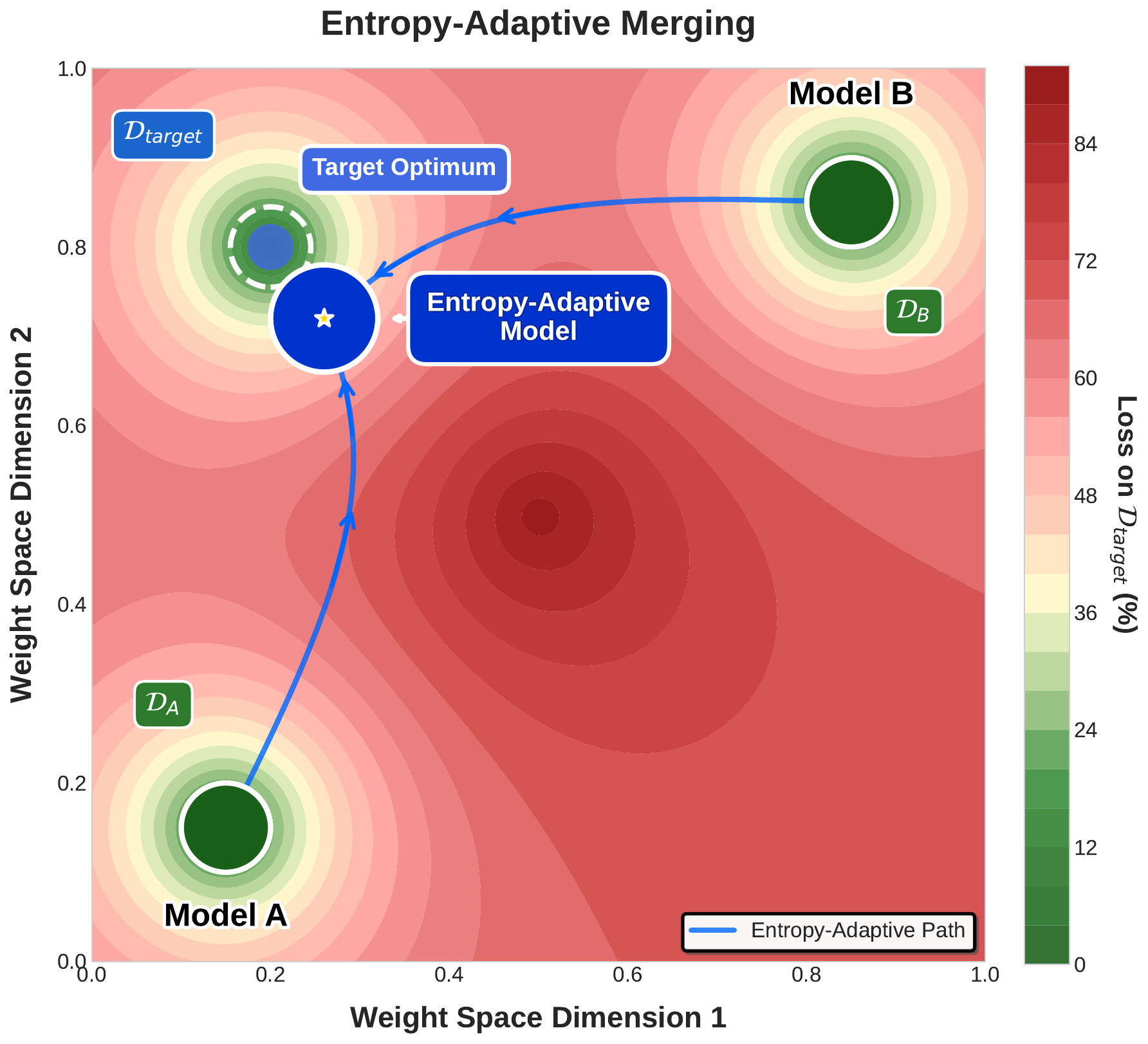}
\end{subfigure}

\caption{
{Illustration of the loss landscape.}
{(a)} Mean merging: The linear path between models crosses a high-loss barrier because straightforward mean connectivity assumes compatible representations across all the layers.
{(b)} Our method learns adaptive merging coefficients that follow a lower-loss path, resolving feature incompatibility near the unseen target optimum.
}

\label{fig:weight_spaces}

\vspace{-2mm}
\end{figure}

A standard baseline \lflf{for model merging} is to use mean merging~\citep{izmailovAveragingWeightsLeads2019,Modelsoups_ICML2022,kaddour2022flat,rame2022diverse}:
\begin{equation}
    \btheta_{\text{mean}} = \sum_{k=1}^K \alpha_k \btheta_s^{(k)} \qquad
W_{\bphi,\text{mean}} = \sum_{k=1}^K \alpha_k W_{\bphi}^{(k)}\,\lflf{,} \label{eq:mean_merge}
\end{equation}
where $\balpha \in \Delta^{K-1}$ with
$\Delta^{K-1}$ \lf{denotes} the set of valid weights over the $K$ source models (i.e., non\lflf{-}negative coefficients that sum to one). 
Intuitively, mean-merging-based methods work well when the source models are trained on the same subsets of training data~\cite{frankle2020linear}, so that interpolating their parameters leads to a smoother parameter space for the merged model. 
This assumption is closely related to linear mode connectivity: if two solutions are connected by a low-loss linear path \lflf{in the loss landscape}, then simple averaging is more likely to remain in a low-loss region~\cite{frankle2020linear}. However, our setup involves weights of models trained on non-overlapping heterogeneous domains, \lflf{which prevents the existence of a low-loss linear path. As illustrated in Fig.~\ref{fig:weight_spaces}a, this results in the mean model potentially ending up in a high loss region. }

In the following paragraphs, for mean merging, we detail three representative failure modes: (A) cross-domain parameter misalignment, (B) encoder-head mismatch, and (C) \lflf{m}isalignment of layers for online shifts.

\textbf{(A) Data diversity (cross-domain heterogeneity) makes linear mode connectivity harder.} Linear mode connectivity studies whether two or multiple trained solutions can be connected by a low-loss {linear} path in weight space~\cite{frankle2020linear}.
During source training, each domain-specific model consists of an encoder and a classification head, with its model parameters denoted by 
$\boldsymbol{f}$ as in Eq.~(\ref{eq:expert_form}), and the $k$-th domain-specific model as $\boldsymbol{f}_s^{(k)}$.
Given two domain-specific models $i$ and $j$, mean merging performs linear interpolation using
\(
\boldsymbol{f}(\lambda) = (1-\lambda)\boldsymbol{f}_s^{(i)} + \lambda \boldsymbol{f}_s^{(j)} ~ \text{with} ~ \lambda\in[0,1],
\)
and chooses a fixed point on this segment (e.g., $\lambda=\tfrac{1}{2}$ for equal weights \lflf{(Fig~\ref{fig:weight_spaces}a)}, or more generally $\lambda$ induced by $\balpha$ in Eq.~(\ref{eq:mean_merge}). Even when both endpoints $\boldsymbol{f}_s^{(i)}$ and $\boldsymbol{f}_s^{(j)}$ are good solutions independently, the straight line segment $(1-\lambda)\boldsymbol{f}_s^{(i)}+\lambda\boldsymbol{f}_s^{(j)}$ may pass through high-loss regions~\cite{garipovLossSurfacesMode2018}.
With cross-domain heterogeneity, mean merging is often less reliable because the models may encode different features and decision boundaries.

To make this misalignment explicit, Fig.~\ref{fig2:heterogeneity_analysis}a reports two complementary {layer-wise} measures between a pair of source models $(i,j)$. Specifically, let $\mathbf{p}^{(k)}_\ell$ be the flattened parameter vector of layer $\ell$ (encoder block or classification head) in model $k$. We measure {directional drift} between $\mathbf{p}^{(i)}_\ell$ and $\mathbf{p}^{(j)}_\ell$ by the angle (in degrees), and {scale drift} by the expected norm ratio $\mathbb{E}\left[|\mathbf{p}^{(i)}_\ell| / |\mathbf{p}^{(j)}_\ell|\right]$.
Linear interpolation is sensitive to both: large angular drift indicates incompatible directions in weight space, while large-scale drift means averaging can be dominated by the higher-norm parameter. 
\lflf{Overall, we observe in Fig.~\ref{fig2:heterogeneity_analysis}a non-collinear and scale-inconsistent drift across the layers that increases with depth.} 
\lflf{This indicates that a single global mean coefficient (Eq.~\ref{eq:mean_merge}) can be appropriate for some layers yet harmful for others across different domain-specific models.}

\textbf{(B) Encoder and classification head mismatch: progressive misalignment across depth.}
Each domain-specific model $k$ decomposes into an encoder and a classification head, $(\btheta_s^{(k)}, W_{\bphi}^{(k)})$.
In our setup, all source models are initialized with the same pretrained encoder (e.g., ImageNet) and fine-tuned on their respective domains, whereas the classification head has to be trained from scratch.
This creates an asymmetry: early encoder blocks can remain relatively compatible across domains, while the head can drift due to domain-specific class priors and decision boundaries.

Fig.~\ref{fig2:heterogeneity_analysis}b \lflf{shows} this effect using {mean angular drift} across depth.
{Concretely, for each depth $\ell$ (encoder blocks $\ell=0,\dots,11$) and for the classification head, we compute the pairwise directional drift angles between all model pairs $(i,j)$ using the flattened vectors $\mathbf{p}^{(i)}_\ell$ and $\mathbf{p}^{(j)}_\ell$, and then average these angles across all ${K}/{2}$ pairs to obtain a single mean value per depth (in degrees).}
The resulting curve shows a clear depth effect: drift is smaller in shallow encoder blocks, increases in deeper blocks, and exhibits a pronounced jump at the classification head.
{This shows that, unlike the pairwise drift in Fig.~\ref{fig2:heterogeneity_analysis}a, mean angular drift isolates a {systematic} depth-wise trend shared across domains, indicating that encoder and head can diverge at different rates and therefore should not be forced to share the same merging behavior.}
Thus, even when encoder blocks remain moderately aligned, classification heads can be highly misaligned, further motivating separate treatment of the encoder and the heads during merging.

\textbf{(C) Fixed mean merging is misaligned and prone to failure for online realistic medical shifts.} At test time for realistic scenarios, the target stream is typically non-i.i.d., and each batch may arrive from a time-varying mixture $\mathcal{D}_{t,n}$=$\sum_{m=1}^M \pi_{n,m}\mathcal{D}_t^{(m)}$ with time-dependent weights $\pi_n$. In these cases, mean merging cannot adaptively respond to such shifts. 
As shown in Fig.~\ref{fig2:heterogeneity_analysis}c mean merging is even structurally prone to fail for heterogeneous data, as misaligned parameters inherently attenuate their magnitudes.

Specifically, when two model parameter vectors form an angle $\vartheta$, their equal-weight average shrinks in norm by a factor $\cos(\vartheta/2)$, yielding a corresponding {signal loss from averaging} of $L_{\text{merge}}(\vartheta)=100\,(1-\cos(\vartheta/2))\%$ (Fig.~\ref{fig2:heterogeneity_analysis}c). 
Using the depth-dependent angular drift measured in Fig.~\ref{fig2:heterogeneity_analysis}b, panel (c) plots the implied attenuation across network depth and highlights a critical-loss regime (e.g., $\ge 50\%$).
The trend is strongly depth-dependent. The attenuation is comparatively small for shallow encoder blocks but increases for deeper ones. It becomes most severe at the classification head. This shows that uniform mean averaging across layers can weaken logits during inference.

\subsection{Entropy-adaptive merging}
\label{section:entropy_merging}
Taking the limitations of mean-merging into account,
\lf{we propose entropy-adaptive merging to deploy a single model on an unlabeled, online target stream.} \lf{W}e propose to estimate merging coefficients {online} using only forward-pass statistics on \lf{the target batch} $\mathcal{B}_t$, then instantiate the resulting single merged model for inference per batch.

\lf{For each domain specific model $k$, we compute the entropy
$\,\mathcal{H}_k(\bx;\tau)\triangleq H\!\big(\mathrm{softmax}(\bz_k(\bx)/\tau)\big)\,$, with the 
 logits $\,\bz_k(\bx)=(W_{\bphi}^{(k)})^\top g(\bx;\btheta_s^{(k)})\,$ and $H(\bp)\triangleq-\sum_{c=1}^C p_c\log p_c$. } 
\lf{and compute} the batch score
\begin{equation}
E_{t,k}
\;=\;
\frac{1}{|\mathcal{B}_t|}
\sum_{\bx \in \mathcal{B}_t}
\mathcal{H}_k(\bx;\tau).
\end{equation}

We then map $\{E_{t,k}\}_{k=1}^K$ to the weights via:
\begin{equation}
\label{eq:alpha_inv_entropy}
\tilde{\alpha}_{t,k}
=\frac{(E_{t,k}+\epsilon)^{-1}}{\sum_{j=1}^K (E_{t,j}+\epsilon)^{-1}},
\qquad \tilde{\balpha}_t\in\Delta^{K-1}.
\end{equation}
This rule implements a specific selection mechanism: domain-specific models that are more confident on $\mathcal{B}_t$ (i.e., have lower predictive entropy) receive larger weights in the merged model, while models that are less confident (higher entropy) are automatically down-weighted.

Finally, we instantiate a single merged model for every \lf{target batch t}:
\begin{equation}
\label{eq:merged_model_t}
\btheta_t=\sum_{k=1}^K \tilde{\alpha}_{t,k}\btheta_s^{(k)}\,,
\qquad
W_{\bphi,t}=\sum_{k=1}^K \tilde{\alpha}_{t,k}W_{\bphi}^{(k)}\;.
\end{equation}
Predictions are then produced by a feed-forward pass:
\begin{equation}
\label{eq:predict_t}
f_t(\bx)=(W_{\bphi,t})^\top g(\bx;\btheta_t)\,,\qquad \bx\in\mathcal{B}_t\;.
\end{equation}
Importantly, computing $\tilde{\balpha}_t$ requires only forward evaluation of the $K$ domain-specific models on $\mathcal{B}_t$ and does not require gradients, labels, or storing target data beyond the current batch.

In the following section, building on entropy-adaptive merging, we introduce three architecturally informed changes that refine merging under heterogeneous, non-stationary shifts.

\subsection{Decoupling encoder and classification head coefficients}

Standard model merging typically applies a single coefficient vector $\tilde{\balpha}_t$ to all parameters, coupling the encoder and the classification head; this can be suboptimal when the head exhibits the connectivity limitations detailed above.
Recent adaptive and layer-wise methods, including AdaMerging~\cite{AdaMerging_ICLR2024}, layer-wise merging~\cite{adilova2024layerwise}, relax this coupling by learning different coefficients across tasks and/or layers.
In contrast, our work first demonstrates classification head-specific failure under domain shift and therefore proposes to merge with {module-decoupled} coefficients. 

More specifically, to address the encoder and classification head mismatch \lf{(See Fig~\ref{fig2:heterogeneity_analysis}b)}, we propose to use {decoupled} merging coefficients:
\begin{equation}
\label{eq:decoupled}
\btheta_t=\sum_{k=1}^K \alpha^{(\mathrm{enc})}_{t,k}\btheta_s^{(k)},
\qquad
W_{\bphi,t}=\sum_{k=1}^K \alpha^{(\mathrm{head})}_{t,k}W_{\bphi}^{(k)}.
\end{equation}

\paragraph{Encoder coefficients.} For the encoder, we set the same merging coefficients $\tilde{\alpha}_{t,k}$, defined via entropy-weighted mixture (Eq.~\ref{eq:alpha_inv_entropy}).

\paragraph{Classification head coefficients.}
For the classification head, we compute coefficients $\{\alpha^{(\mathrm{head})}_{t,k}\}_{k=1}^K$ separately to avoid destructive averaging when heads are strongly misaligned. 
Given the unlabeled target batch $\mathcal{B}_t$, we select a {well-performing batch-selected model} $k_t^\star$ using a reliability score that combines low predictive entropy and high data augmentation consistency:
\begin{equation}
\label{eq:head_expert}
k_t^\star \;=\; \arg\max_{k\in\{1,\dots,K\}}
\Big(\tfrac{1}{E_{t,k}}\Big)\big(1+C_{t,k}\big),
\end{equation}
where $K$ is the number of domain-specific models, $E_{t,k}$ is the predictive entropy of domain-specific model $k$ on $\mathcal{B}_t$ (lower means more confident), and $C_{t,k}$ the augmentation consistency. 
The augmentation consistency $C_{t,k}$ quantifies how invariant the domain-specific model $k$'s predicted class probabilities are on $\mathcal{B}_t$ under label-preserving simple augmentations (higher means more stable predictions across augmented views).

We then calculate the head coefficients by a normalized entropy-gap weighting:
\begin{equation}
\label{eq:head_coeffs_single}
\alpha^{(\mathrm{head})}_{t,k}
\;=\;
\frac{\exp\!\big(-\tau\,|E_{t,k}-E_{t,k_t^\star}|\big)}
{\sum_{j=1}^{K}\exp\!\big(-\tau\,|E_{t,j}-E_{t,k_t^\star}|\big)},
\end{equation}
where $\alpha^{(\mathrm{head})}_{t,k}\ge 0$ and $\sum_{k=1}^K \alpha^{(\mathrm{head})}_{t,k}=1$ by construction (softmax-based normalization), and $\tau>0$ is a temperature controlling sharpness (larger $\tau$ concentrates more weight on experts whose entropy is closest to $E_{t,k_t^\star}$, smaller $\tau$ spreads weights more evenly). Additionally, we apply an exponential moving average to improve stability. More details are provided in the Appendix.

\textbf{Algorithms. } An algorithm describing test-time online merging is depicted below (Algorithm~\ref{alg:test_merge_short})  and for source training in the Appendix Section~\ref{appendix:implementational}.

\begin{algorithm}[ht!]
\small
\caption{\textbf{At Test-time: forward-only} \\
{\textbf{Input:}} Frozen domain specific models $\{(\btheta_s^{(k)},W_{\bphi}^{(k)})\}_{k=1}^K$; unlabeled target batches $\{\mathcal{B}_t\}_{t=1}^T$; $\tau,\epsilon$; EMA rate $\mu$; head option: {shared} (Eq.~\ref{eq:merged_model_t}) or {entropy-gap} (Eq.~\ref{eq:head_coeffs_single}) \\
\textbf{Output:} Predictions $\{\hat{\by}_t\}_{t=1}^T$.
}
\label{alg:test_merge_short}
\begin{algorithmic}[1]
\STATE Initialize $\balpha_0 \gets (1/K,\ldots,1/K)$.
\FOR{$t=1$ \textbf{to} $T$}
    \STATE Compute $\{E_{t,k}\}_{k=1}^K$ on $\mathcal{B}_t$.
    \STATE Compute $\tilde{\balpha}_t$ by inverse-entropy normalization (with $\epsilon$).
    \STATE Compute $\{C_{t,k}\}_{k=1}^K$ on $\mathcal{B}_t$ (augmentation-consistency).
    \STATE Select head expert $k_t^\star$ (Eq.~\ref{eq:head_expert}) and compute head coefficients
    $\{\alpha^{(\mathrm{head})}_{t,k}\}_{k=1}^K$ (Eq.~\ref{eq:head_coeffs_single}).
    \STATE Merge to form $(\btheta_t,W_{\bphi,t})$ with decoupled weights and new co-efficients:\\
    $\btheta_t \gets \sum_{k=1}^K \alpha_{t,k}\btheta_s^{(k)}$,\;
    $W_{\bphi,t} \gets \sum_{k=1}^K \alpha^{(\mathrm{head})}_{t,k} W_{\bphi}^{(k)}$ (Eq.~\ref{eq:decoupled}, Eq.~\ref{eq:head_coeffs_single}).
    \STATE Apply exponential moving average and Predict on $\mathcal{B}_t$ with the merged model (Eq.~\ref{eq:predict_t}).
\ENDFOR
\end{algorithmic}
\end{algorithm}

\begin{table*}[t]
    \centering
 \caption{\textbf{State-of-the-art comparisons across medical datasets}. We report the mean accuracy for ViT-B/32 (B32) and ViT-B/16 (B16) under the standard {leave-one-domain-out}~\cite{gulrajani2020search} setting, where domain-specific models are trained on respective domains and evaluated on an unseen target domain (adaptively). We re-implement the baselines as either offline or online model-merging methods and detail them in the Appendix. Our entropy-adaptive method achieves the best performance (\textbf{bold}).}
    \label{tab:medical_results}
    \small
    \setlength{\tabcolsep}{6pt}

    \begin{tabular}{l|cc|cc|cc|cc|cc}
        \toprule
        & \multicolumn{10}{c}{\textbf{\textit{5 Medical Datasets}}} \\
        \cmidrule(lr){2-11}
        & \multicolumn{2}{c|}{\textbf{MiDog}} 
        & \multicolumn{2}{c|}{\textbf{Organs}}
        & \multicolumn{2}{c|}{\textbf{Histo}}
        & \multicolumn{2}{c|}{\textbf{ISIC Skin}}
        & \multicolumn{2}{c}{\textbf{Messidor}} \\
        
        \cmidrule(lr){2-3} \cmidrule(lr){4-5} \cmidrule(lr){6-7} \cmidrule(lr){8-9} \cmidrule(lr){10-11}
        \textbf{Method}
        & \small B32 & \small B16
        & \small B32 & \small B16
        & \small B32 & \small B16
        & \small B32 & \small B16
        & \small B32 & \small B16 \\
        
        \midrule
        \rowcolor{mColor2}
        \multicolumn{11}{l}{\textcolor{black}{\textit{Offline Model merging}}} \\
        
        Ensemble
        & 89.24 %
        & 89.22 %
        & 69.87 %
        & 73.71 %
        & 92.16 %
        & 89.87 %
        & 86.89 %
        & 86.24 %
        & 49.75 %
        & 54.58 %
        \\
        
        DiWA~\cite{rame2022diverse}
        & 89.65 %
        & 89.07 %
        & 66.93 %
        & 71.26 %
        & 89.48 %
        & 89.48 %
        & 85.60 %
        & 85.82 %
        & 49.50 %
        & 52.81 %
        \\
        
        \midrule
        \rowcolor{mColor2}
        \multicolumn{11}{l}{\textcolor{black}{\textit{Online Model merging}}} \\
        
        TIESMerging~\cite{TiesMergingNeurIPS2023}
        & 88.77 %
        & 89.53 %
        & 67.75 %
        & 71.74 %
        & 91.30 %
        & 88.06 %
        & 85.12 %
        & 77.23 %
        & 48.83 %
        & 54.25 %
        \\
        
        Adamerging~\cite{AdaMerging_ICLR2024}
        & 89.39 %
        & 89.25 %
        & 67.50 %
        & 73.36 %
        & 93.56 %
        & 88.95 %
        & 85.87 %
        & 85.54 %
        & 42.22 %
        & 54.58 %
        \\
        
        Task Arithmetic~\cite{TaskArithmetic_ICLR2023}
        & 89.62 %
        & 89.30 %
        & 67.80 %
        & 73.56 %
        & 91.57 %
        & 89.00 %
        & 85.52 %
        & 85.64 %
        & 41.12 %
        & 54.60 %
        \\
        
        RegMean++~\cite{nguyen2025regmeanpp}
        & 88.40 %
        & 88.75 %
        & 65.52 %
        & 69.20 %
        & 92.77 %
        & 88.50 %
        & 85.02 %
        & 86.04 %
        & 47.41 %
        & 51.18 %
        \\
        
        FisherMerging~\cite{FisherMerging_NeurIPS2022}
        & 88.32 %
        & 88.80 %
        & 65.30 %
        & 68.83 %
        & 91.60 %
        & 89.55 %
        & 84.80 %
        & 84.90 %
        & 51.75 %
        & {55.40} %
        \\
        
        \toprule
        \textbf{\textit{Entropy-Adaptive (Ours)}}
        & \textbf{89.70} %
        & \textbf{91.00} %
        & \textbf{70.21} %
        & \textbf{74.60} %
        & \textbf{93.92} %
        & \textbf{91.00} %
        & \textbf{87.10} %
        & \textbf{86.60} %
        & \textbf{52.33} %
        & \textbf{55.73} %
        \\
        
        \bottomrule
    \end{tabular}
\end{table*}

\begin{table*}[t]
    \centering
 \caption{\textbf{State-of-the-art comparisons across natural vision datasets}. We report the mean accuracy for ViT-B/32 (B32) and ViT-B/16 (B16) and follow the same settings as in Table~\ref{tab:medical_results}. We reimplement the baselines as either offline or online model-merging methods and detail them in the Appendix. Our entropy-adaptive method achieves the best performance (\textbf{bold}).}
    \label{tab:natural_results}
    \small
    \setlength{\tabcolsep}{6pt}

    \begin{tabular}{l|cc|cc|cc|cc}
        \toprule
        & \multicolumn{8}{c}{\textbf{\textit{4 Natural Vision Datasets}}} \\
        \cmidrule(lr){2-9}
        & \multicolumn{2}{c|}{\textbf{PACS}} 
        & \multicolumn{2}{c|}{\textbf{VLCS}}
        & \multicolumn{2}{c|}{\textbf{Office-Home}}
        & \multicolumn{2}{c}{\textbf{Terra}} \\
        
        \cmidrule(lr){2-3} \cmidrule(lr){4-5} \cmidrule(lr){6-7} \cmidrule(lr){8-9}
        \textbf{Method}
        & \small B32 & \small B16
        & \small B32 & \small B16
        & \small B32 & \small B16
        & \small B32 & \small B16 \\
        
        \midrule
        \rowcolor{mColor2}
        \multicolumn{9}{l}{\textcolor{black}{\textit{Offline Model merging}}} \\
        
        Ensemble
        & 80.46 & 84.00
        & 78.51 & 77.21
        & 74.20 & 77.83
        & 30.42 & 41.22 \\
        
        DiWA~\cite{rame2022diverse}
        & 83.91 & 84.72
        & 77.54 & 77.67
        & 75.46 & {78.44}
        & 27.64 & 40.43 \\
        
        \midrule
        \rowcolor{mColor2}
        \multicolumn{9}{l}{\textcolor{black}{\textit{Online Model merging}}} \\
        
        TIESMerging~\cite{TiesMergingNeurIPS2023}
        & 82.91 & 84.20
        & 78.30 & 74.30
        & 72.55 & 72.50
        & 23.51 & 36.20 \\
        
        Adamerging~\cite{AdaMerging_ICLR2024}
        & 83.25 & 85.56
        & 78.12 & 78.20
        & 75.17 & 77.90
        & 23.33 & 32.50 \\
        
        Task Arithmetic~\cite{TaskArithmetic_ICLR2023}
        & 81.82 & 85.07
        & 78.12 & 77.90
        & 74.90 & 77.27
        & 22.31 & 41.21 \\
        
        RegMean++~\cite{nguyen2025regmeanpp}
        & 76.52 & 82.15
        & 59.32 & 79.71
        & 68.70 & 76.65
        & 27.90 & 32.50 \\
        
        FisherMerging~\cite{FisherMerging_NeurIPS2022}
        & 82.91 & 82.22
        & 76.67 & 78.10
        & 68.25 & 72.12
        & 27.87 & 32.25 \\
        
        \toprule
        
        \textbf{\textit{Entropy-Adaptive (Ours)}}
        & \textbf{86.15} & \textbf{87.60}
        & \textbf{80.33} & \textbf{79.80}
        & \textbf{76.34} & \textbf{80.20}
        & \textbf{32.15} & \textbf{46.74} \\
        
        \bottomrule
    \end{tabular}
    \vskip -0.3cm
\end{table*}

\section{Experiments}

\textbf{9 Datasets with 2 ViT backbones.} We re-implement state-of-the-art baselines on 9 standard \textit{public} domain-generalization classification datasets for our problem setup, 5 from medical imaging and four from natural vision, using both ViT-B/16 and ViT-B/32.

\noindent\textbf{Medical imaging datasets.}
(i) {MiDog Atypical (MiDog)}~\cite{midog2025atypical}: 454 histopathological images; task: normal vs.\ atypical mitosis classification; domains reflect different tumor types (7). (ii) {Organs}~\cite{woerner2024comprehensive}: 1,645 2D CT slices; task: 11-organ recognition; domains reflect acquisition plane (axial/coronal/sagittal). (iii) {Histopantum (Histo)}~\cite{zamanitajeddin2024benchmarking}: 281,142 histopathology patches; task: tumor vs.\ non-tumor classification; domains reflect different cancer types (4) (iv) {ISIC Skin}~\cite{tschandl2018ham10000}: 10,015 dermoscopy images; task: melanoma classification (1); domains reflect common acquisition artifacts (e.g., hair, ruler, bubbles). (v) {Messidor}~\cite{messidor_ds}: 1,200 retinal fundus images; task: diabetic retinopathy grading (0--3); domains reflect acquisition centers (3).

\noindent\textbf{Natural vision datasets.}
(vi) {PACS}~\cite{li2017deeper}: 9,991 images; 7 classes; domains capture style shifts (Photo/Art/Cartoon/Sketch). (vii) {VLCS}~\cite{fang2013video}: 10,729 images; 5 classes; domains correspond to different source datasets (VOC/LabelMe/Caltech/SUN). (viii) {Office-Home}~\cite{venkateswara2017deep}: 15,500 images; 65 classes; domains capture appearance/context shifts (Art/Clipart/Product/Real). (ix) {TerraIncognita (Terra)}~\cite{beery2018recognition}: camera-trap wildlife recognition; domains reflect camera locations. We provide additional dataset details in the Appendix.

\textbf{Implementation details.} We train the single-source domain-specific models using the DiWA codebase~\cite{rame2022diverse}. For two Vision Transformer (ViT) backbones~\cite{dosovitskiy2020image}, we train multiple instances under different hyperparameter settings with standard cross-entropy classification as in DiWA. Each domain-specific model is trained only on its own source domain, and we select the best checkpoint using in-domain validation. At test time, we assume access to \(K\) frozen domain-specific models and, at each time step, merge them into a single predictor and evaluate on the current target batch using a forward-only pass (no backpropagation and no target-time updates). We report accuracy over the target stream. All experiments are run on NVIDIA A100 GPUs; our method is non-parametric and incurs only lightweight per-batch overhead. For all ablations, we use ViT-B/32 as a backbone and a batch size of 32. We provide additional details in the Appendix and will release the code publicly.

\textbf{State-of-the-art comparisons.} Table~\ref{tab:medical_results} reports medical-domain results for both ViT-B/32 and ViT-B/16 under the standard leave-one-domain-out domain generalization protocol~\cite{gulrajani2020search}, using both offline and online merging with a test-time batch size of 32 samples. Across all reported medical datasets, \textit{entropy-adaptive (Ours)} achieves the best accuracy for both backbones, outperforming fixed merging (DiWA~\cite{rame2022diverse}) and strong online baselines such as TIES-Merging~\cite{TiesMergingNeurIPS2023}, AdaMerging~\cite{AdaMerging_ICLR2024}, which uses backpropagation, RegMean++~\cite{nguyen2025regmeanpp}, and FisherMerging~\cite{FisherMerging_NeurIPS2022}. The gains are especially clear in domains with stronger domain heterogeneity (e.g., Organs dataset~\cite{woerner2024comprehensive}), while in already saturated settings (e.g., MiDog Atypical~\cite{midog2025atypical}), improvements are modest but remain consistently positive. Our performance gains stem from entropy-adaptive, per-batch coefficient estimation for domain-specific models, and from decoupling the encoder and classification head, to handle their distinct cross-domain misalignment. 

Table~\ref{tab:natural_results} shows the corresponding results on natural-image datasets, again for both ViT-B/32 and ViT-B/16. Our method remains best on all datasets and both backbones, with particularly large improvements on the harder shift benchmarks (VLCS and TerraIncognita), where naive averaging and several online merging baselines can suffer from negative transfer. These results reinforce that entropy-adaptive, per-batch coefficient estimation with decoupled encoder and classification head weighting generalizes beyond the medical setting and remains effective across architectures.

\subsection{Additional experiments and Benefits}
\begin{figure}[t]
\centering
\includegraphics[width=1.0\linewidth, height=1.0\textheight, keepaspectratio]{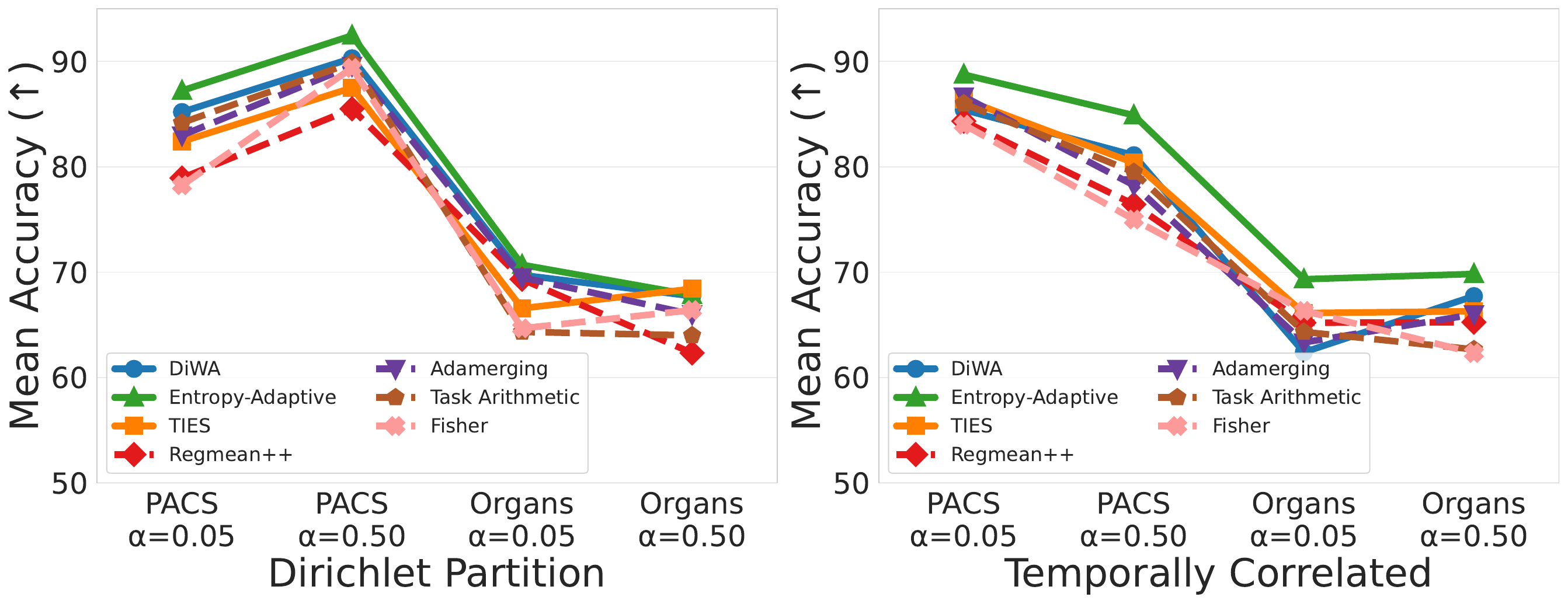}

\caption{\textbf{\textbf{Consistent improvements under Dirichlet and Temporal sampling. }} Mean accuracy (\%) on PACS and Organs under Dirichlet splits ({0.05, 0.50}) and Temporally correlated data. Our Entropy-Adaptive (green) consistently outperforms the baselines with larger gains under a stronger skew factor (\(\alpha=0.05\)).}

\label{fig:dirichlet}
\vspace{-4mm}
\end{figure}

\textbf{Consistent improvements under Dirichlet and Temporal sampling. }
We also evaluate our method under severe shifts, where each incoming batch is formed via a Dirichlet-class partition~\cite{gong2022note} and temporally correlated, yielding non-i.i.d.\ target streams with strong class imbalance. {Figure~\ref{fig:dirichlet}a shows the mean accuracy on PACS and Organs under Dirichlet splits and temporally correlated target streams.}
Across both PACS and Organs, our entropy-adaptive merging achieves consistently higher mean accuracy than all the baselines, with larger gains under the stronger skew factor ($\alpha=0.05$). Our entropy-adaptive merging performs well even under strong shifts because it adapts the model to each batch using target information, suppressing and amplifying the required domain-specific models.

\begin{table}[t]
\centering

\resizebox{0.95\columnwidth}{!}{
\begin{tabular}{lccll}
\toprule
& \multicolumn{2}{c}{\textbf{Coefficients strategy}} &  \multicolumn{2}{c}{\textbf{Mean accuracy}} \\
\cmidrule(lr){2-3} \cmidrule(lr){4-5}
\textbf{Method} & Entropy only (Eq.~\ref{eq:alpha_inv_entropy}) & Decoupled head coef (Eq.~\ref{eq:head_coeffs_single}) & \textbf{PACS} & \textbf{Organs}\\
 \midrule
 \midrule
\multirow{3}*{\textit{\textbf{Entropy-Adaptive}}} & \ymark &  & {85.23} \scriptsize{$\pm$0.2} &  69.81 \scriptsize{$\pm$0.2}\\
 & & \ymark & 84.27 \scriptsize{$\pm$0.2}& 67.80 \scriptsize{$\pm$0.3} \\
 & \ymark & \ymark &  \textbf{87.33} \scriptsize{$\pm$0.2} &  \textbf{70.70} \scriptsize{$\pm$0.2}\\
\bottomrule
\end{tabular}}
\vspace{2mm}
\caption{\textbf{Benefits of Entropy only and decoupled head coefficients} for ViT-B/32 on {PACS} and {Organs}. Both components improve performance; together, they deliver the strongest gains, highlighting the need for distinct coefficients for the encoder and the classification head.}
\label{tab:ablation}
\vspace{-6mm}
\end{table}

\textbf{Benefits of our coefficients for merging. }
Our method yields a batch-specific merged model through entropy-adaptive coefficients and decoupled encoder and classification head weights. To show the benefits of our contributions, we demonstrate them on both the PACS and Organs datasets using ViT-B/32 (Table~\ref{tab:ablation}) under Dirichlet sampling ($\alpha{=}0.05$). This sampling strategy creates highly skewed non-i.i.d.\ batches (29/32 samples from one class on PACS with a total of 7 classes). In this challenging setting, entropy-adaptive merging alone is already strong (second-best), while adding decoupled coefficients achieves additional gains, consistent with the encoder and classification head exhibiting distinct cross-domain misalignment. Overall, the two components are complementary and beneficial when the target-stream imbalance is severe.

\noindent\textbf{Compute and time analysis.} Our method is forward-only at test time (no backpropagation or optimizer). Per batch, it runs the $K$ frozen models to compute entropies and then forms a merged model for inference. On PACS, this takes 40 seconds on a single Nvidia A100 GPU and 3.5GB to use the $K$ domain-specific models, comparable to mean merging~\cite{Modelsoups_ICML2022,rame2022diverse}.

We also provide additional discussions in Appendix Sec.~\ref{Appendix:Discussions}. %

\section{%
Conclusion \lff{and Outlook}}

In this work, we propose an {entropy-adaptive merging} method that obtains a batch-specific, inference-time-merged model for each unlabeled target batch using only forward passes.
We empirically demonstrate that uniform mean merging is prone to failure under heterogeneous cross-domain shifts, and address this by selecting target-specific coefficients online without modifying domain-specific model parameters.
To prevent degradation of classification head weights, we decouple the encoder and classifier-head coefficients, combining smooth encoder mixtures with a more conservative head rule, and stabilize the online trajectory with an exponential moving average.
Across diverse medical and natural datasets, including challenging scenarios, this strategy consistently improves robustness while preserving the operational simplicity of a single merged model. We consider reliance on entropy as a confidence signal under severe out-of-distribution conditions a limitation. However, this may be mitigated by strengthening the confidence estimate through consistency signals beyond entropy alone.

\section*{Acknowledgments}
This work is supported by DAAD programme Konrad Zuse Schools of Excellence in Artificial Intelligence and the Munich Center for Machine Learning, both sponsored by the Federal Ministry of Research, Technology and Space. DML and JAS received funding from HELMHOLTZ IMAGING, a platform of the Helmholtz Information and Data Science Incubator. 

\nocite{ambekar2022skdcgn}
\nocite{ambekarvariational}
\bibliography{example_paper}
\bibliographystyle{icml2026}

\newpage
\clearpage
\appendix

\appendix
\begin{leftline}
    
{
\LARGE{\textsc{Appendix}}
}
\end{leftline}

\section{Additional Datasets details}
\noindent\textbf{Natural Vision datasets.}
\lfs{We use four natural vision datasets, each containing multiple visually distinct \emph{domains} (e.g., acquisition source or artistic style) with a shared label space, enabling leave-one-domain-out evaluation.} 
(i) \textit{PACS}~\cite{li2017deeper} contains $9,991$ images from \lff{seven} classes across four domains (Photo, Art-painting, Cartoon, Sketch), and is widely used to study style shifts that alter low-level statistics while preserving semantic labels. 
(ii) \textit{VLCS}~\cite{fang2013video} contains $10,729$ images from \lff{five} classes across four domains (VOC2007, LabelMe, Caltech, SUN), capturing dataset-source shifts from different collection pipelines and scene/object statistics. 
(iii) \textit{Office-Home}~\cite{venkateswara2017deep} contains $15,500$ images spanning $65$ classes over four domains (Art, Clipart, Product, Real-World), providing a larger and more fine-grained DG benchmark. 
(iv) \textit{TerraIncognita}~\cite{beery2018recognition} is a camera-trap wildlife recognition dataset where domain shift is induced by geographically distinct camera locations and environmental conditions, with the common preprocessing that treats locations as domains.

\noindent\textbf{Medical imaging datasets. }
To evaluate robustness under clinically realistic distribution shifts, we consider medical imaging benchmarks spanning histopathology, dermoscopy, mammography, and CT anatomy classification. (v) \textit{MiDog Atypical} (MIDOG 2025 Atypical Training Set)~\cite{midog2025atypical} contains 11,939 mitotic figures from 454 labeled images across seven domains (breast carcinoma, lung carcinoma, lymphosarcoma, neuroendocrine tumor, cutaneous mast cell tumor, melanoma, and soft tissue sarcoma). 
The dataset is explicitly constructed to induce domain shift via tumor-type heterogeneity, species variation (human and canine), scanner diversity (five different whole-slide scanners), and lab-specific tissue preparation protocols.
Annotations are obtained via a three-expert blinded majority vote for binary classification (normal vs.\ atypical mitosis), making this a challenging DG setting where staining/scanner artifacts and morphology changes can strongly perturb low-level texture statistics while preserving the underlying diagnostic task. (vi) \textit{Organs} is derived from the Liver Tumor Segmentation Benchmark (LiTS) and provides 2D CT slices for multiclass organ recognition under viewpoint changes.
Specifically, we use three variants, {organs\_axial}, {organs\_coronal}, and {organs\_sagittal} each containing 1,645 images with 11 organ labels (heart, left/right lung, liver, spleen, pancreas, left/right kidney, bladder, left/right femoral head). 
This setup naturally supports a DG protocol by treating the slicing plane (axial/coronal/sagittal) as the domain, while keeping the label space fixed across domains. (vii) \textit{HISTOPANTUM} is a pan-cancer tumor detection dataset released in the referenced study, covering four cancer types as domains: colorectal (CRC), uterus (UCEC), ovary (OV), and stomach (STAD), with 40 WSIs per domain sourced from TCGA studies. 
An experienced pathologist annotates tumor and non-tumor regions, from which the dataset is formed as 281,142 patches of size 512$\times$512 (at about 0.5 mpp) that are resized to 224$\times$224 during training and evaluation, with binary labels tumor vs.\ non-tumor. 
This dataset exhibits multiple distribution shifts simultaneously, including covariate shift (center/scanner/stain variability), prior shift (different class proportions across cancer types), and class-conditional shift (tumor morphology differs by cancer type while non-tumor tissue is more consistent). (viii) \textit{ISIC Skin} is instantiated as DG for melanoma classification using ISIC2019 for training/validation (melanoma vs.\ benign), with 12,360 training images and 2,060 validation images. 
Domain labels are defined by artifact annotations, splitting the ISIC2019 training set into five groups: dark corner (2,351), hair (4,884), gel bubble (1,640), ruler (672), and clean (2,796). 
Generalization is evaluated on four OOD test sets: Derm7pt-Dermoscopic (872), Derm7pt-Clinical (839), PH2 (200), and PAD-UFES-20 (531), where the shift includes both artifact-driven changes and modality differences (dermoscopic vs.\ clinical photographs). (ix) \textit{Mammo} (mammography) uses a CBIS-DDSM-style setup where labels include benign, benign-without-callback, and malignant, and evaluation is commonly performed either as a 3-way classification task or as benign-vs-malignant depending on the protocol.

\section{Additional Related work}

\textbf{Test-time adaptation.}
Test-time adaptation optimizes the source trained model during inference target data to address unseen distribution shifts via gradient updates~\cite{chen2022ost,choi2021test,min2023meta,xiao2022learning,sahoo2024layer,xiao2024beyond,liang2023comprehensive,xiao2024any,xiao2023energy,boudiaf2022parameter,kim2025testtime,vray2025reservoirtta,ambekar2025gf,Yang_2024_CVPR,lei2025ttvd,du2024unitta}. The standard fine-tuning approaches typically adjust a fixed set of parameters, such as batch normalization statistics~\cite{zhang2021adaptive,nado2020evaluating}, full model parameters~\cite{wang2021tent,liang2020we}, or single-dimensional linear layers~\cite{jang2022test,iwasawa2021test,zhang2023adanpc}. {Beyond these, recent works have explored overall layer selection~\cite{sahoo2024layer} and surgical fine-tuning~\cite{lee2022surgical} for targeted adaptation.} These methods have been evaluated on unseen scenarios, including noise-corrupted datasets~\cite{wang2021tent,goyal2022test} and datasets with varying domain information~\cite{iwasawa2021test,liang2020we}. Test-time adaptation is orthogonal to model merging: merging decides which single model to deploy by consolidating multiple hospital experts into a single merged model, while Test-time adaptation decides how this merged model can adapt further to the incoming target stream.

\textbf{Federated learning.} In Federated Learning (FL) \cite{mcmahan2017fl}, multiple clients train a global model under the supervision of a central server without sharing their raw data \cite{kairouz2019advances}. The clients only share the local models with the server, which, in turn, performs model aggregation (merging). The federated training is an iterative process that continues for a specified number of communication rounds or until the global model converges. There are different model merging methods in FL including \textit{FedAvg} \cite{mcmahan2017fl}, \textit{FedNova} \cite{wang2020fednova}, and \textit{FedFisher} \cite{jhunjhunwala24fedfisher}.
FedAvg~\cite{mcmahan2017fl} is the de facto standard merging method in FL, which takes the weighted average, based on the sample sizes of the clients, over the weights of the local models. FedNova normalizes the gradients from the clients' local models before merging them. FedFisher computes the Fisher information matrices for the local models and uses them as the basis for the global model. However, FL merging rules target iterative training-time aggregation to a single global model and therefore do not suit our inference setting, which requires online merging of fixed domain-specific models using only unlabeled target batches and forward-only computation.

\noindent\textbf{Mode connectivity.} Mode connectivity provides theoretical grounding for model merging by establishing that independently trained models can often be connected by low-loss paths in weight space~\cite{garipovLossSurfacesMode2018,frankleLinearModeConnectivity2020,draxlerEssentiallyNoBarriers2019}. Here, weight space means the (very high-dimensional) space of all network parameters, that is all weights and biases stacked into one long vector, so each trained model corresponds to a single point in that space. Linear mode connectivity~\cite{frankleLinearModeConnectivity2020} shows that models trained from the same initialization often exhibit approximately linear connectivity, although this can break under strong heterogeneity.
Recent work explores mode connectivity in broader contexts, including loss surface simplexes~\cite{bentonLossSurfaceSimplexes2021}, convexity properties~\cite{yunisConvexityLinearMode2022}, geometric perspectives~\cite{simsek2021geometry}, and proving connectivity via optimal transport~\cite{ferbach2024proving}. 

\section{Additional Implementation details}
\label{appendix:implementational}
\subsection{Source training: Independently training the source models } 

To construct diverse source models, we follow the DIWA method (Diverse Weight Averaging)~\cite{rame2022diverse} setup and train multiple runs per hospital by varying hyperparameters and training procedures, explicitly targeting functional diversity among candidate solutions. 
More specifically, for each domain or hospital $k$ we train a set of candidates $\{f_{k,m}\}_{m=1}^{M_k}$ using standard cross-entropy loss on $(\bx_s^{(k)},\by_s^{(k)})$, and then select one representative \emph{domain specific model} per hospital via the hospital-specific validation split, yielding the final domain specific model pool $\{f_k\}_{k=1}^K$. 

\subsection{Evaluation setting} 
We use a unified online test-time adaptation (TTA) evaluation protocol, where target samples arrive sequentially as unlabeled batches $\{\mathcal{B}_t\}_{t=1}^T$. 
All methods are evaluated under identical data ordering and batching, and we reimplement the standard baselines across all datasets. 
To isolate the effect of \emph{adaptive merging} from general-purpose test-time training, we restrict test-time updates to merging-related parameters (e.g., coefficients/weights that determine how domain-specific models are combined), while keeping the underlying expert backbones fixed. 
For adaptive baselines that rely on optimization, we enforce a fixed compute budget by using the same number of update steps per batch across methods, whereas our approach performs \emph{forward-only} inference and does not backpropagate through target data. 
Operationally, at each time step $t$, the method produces a single merged predictor tailored to $\mathcal{B}_t$ and then evaluates it on $\mathcal{B}_t$ in a feed-forward manner, which matches the deployment constraint of streaming inference with limited compute and no target-time training. 

\subsection{Source training algorithm}

\begin{algorithm}[ht!]
\small
\caption{\textbf{Source training and domain-specific model selection} \\
{\textbf{Input:}} Following~\citet{rame2022diverse},  domains $\mathcal{S}=\{D_s^{(k)}\}_{k=1}^K$ with labeled splits $(\mathcal{D}^{(k)}_{\mathrm{tr}},\mathcal{D}^{(k)}_{\mathrm{val}})$; hyperparameter set $\mathcal{H}$. \\
\textbf{Output:} One domain specific model per domain $\{(\btheta_s^{(k)},W_{\bphi}^{(k)})\}_{k=1}^K$.
}
\label{alg:source_diwa_short}
\begin{algorithmic}[1]
\FOR{$k=1$ \textbf{to} $K$}
    \FOR{each $h\in\mathcal{H}$}
        \STATE Train candidate $(\btheta_{k,h},W_{\bphi,k,h})$ on $\mathcal{D}^{(k)}_{\mathrm{tr}}$ using Eq.~\ref{eq:expert_form}.
        \STATE Compute validation loss $R^{(k)}_{\mathrm{val}}(h)$ on $\mathcal{D}^{(k)}_{\mathrm{val}}$.
    \ENDFOR
    \STATE Select $h_k^\star=\arg\min_{h\in\mathcal{H}} R^{(k)}_{\mathrm{val}}(h)$.
    \STATE Set $(\btheta_s^{(k)},W_{\bphi}^{(k)}) \gets (\btheta_{k,h_k^\star},W_{\bphi,k,h_k^\star})$.
\ENDFOR
\end{algorithmic}
\end{algorithm}

\subsection{Hyperparameters}

We re-implement the following baseline methods for our merging setup using ~\cite{tang2024fusionbench}: RegMean++~\cite{nguyen2025regmeanpp}, AdaMerging~\cite{AdaMerging_ICLR2024}, Task Arithmetic~\cite{TaskArithmetic_ICLR2023}, TIES-Merging~\cite{TiesMergingNeurIPS2023}, and Fisher Merging~\cite{FisherMerging_NeurIPS2022}. Unless stated otherwise, we use the hyperparameter settings recommended in the original papers or official implementations. We will release the code. 

\noindent \textbf{RegMean++.} We follow the default configuration and use 50 batches with a batch size of 32. The Gram-matrix regularization strength is set to $\lambda_{\text{reg}}=0.01$, and we do not downweight off-diagonal entries in the Gram matrix, i.e., {reduce the nondiagonal ratio} is set to 1.0.

\noindent \textbf{AdaMerging.} Task-wise merge coefficients are optimized using Adam with a learning rate of $10^{-3}$ for a single epoch. Coefficients are initialized to 0.3 and updated using a batch size of 32. During optimization, coefficients are clamped to the interval $[0,1]$.

\noindent \textbf{Task Arithmetic.} We use a fixed task-vector scaling coefficient and set the scaling factor $\lambda$ to 0.3 for all experiments.

\noindent \textbf{TIES-Merging.} We adopt the standard configuration with a scaling factor of 0.3 and a trimming ratio $k_{\text{threshold}}=0.2$. The merge operator is chosen as the mean.

\noindent \textbf{Fisher Merging.} Fisher information is estimated using 500 samples. We normalize the Fisher weights, set the minimum Fisher weight to $10^{-6}$ to avoid numerical issues, and use a batch size of 32 throughout.

\begin{figure}[t]
\centering
\includegraphics[width=1.0\linewidth, height=1.0\textheight, keepaspectratio]{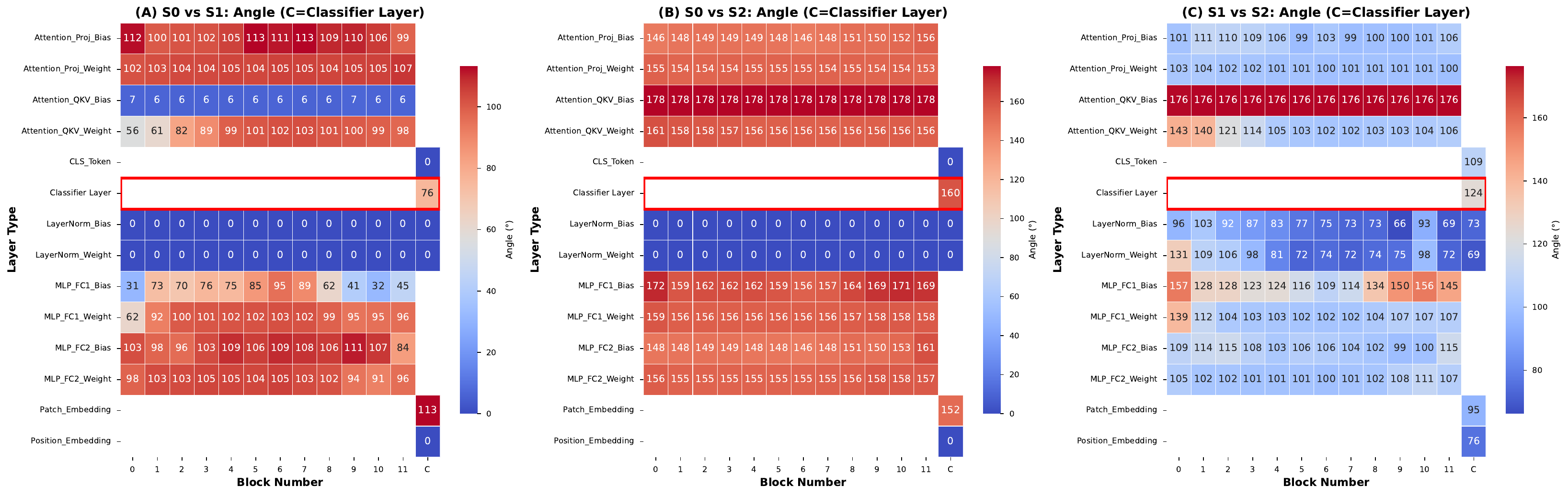}
\caption{\textbf{Layer-wise domain-specific models misalignment.}
Pairwise parameter angles between domain-specific models across ViT and classification head components show that encoder layers are largely aligned, while the classification head (C) is strongly misaligned, motivating head-aware merging.
}

\label{fig:angles_heatmap}
\vspace{-4mm}
\end{figure}

\textbf{Layer-wise domain-specific models misalignment.} Figure~\ref{fig:angles_heatmap} provides a mechanistic ablation supporting our merging design on the PACS dataset using ViT-B32, where S0, S1, and S2 from the figure represent the source models trained on Photo, Cartoon, and Sketch, respectively.  Specifically, we compute pairwise parameter angles as in ~\cite{jang2024model} between domain-specific models and observe a clear layer-wise pattern: representation layers remain comparatively aligned, whereas the classification head is highly misaligned across domains. This explains why global mean merging can be partially effective (when it averages aligned encoder components) yet fails sharply when applied uniformly to the head, which lacks a shared basin across hospitals. 
Motivated by this, our method decouples encoder and classification head merging, enabling soft entropy-weighted fusion for the encoder while using a more conservative rule for the head during online inference.

\section{Additional Discussions: Rationales and Motivation}
\label{Appendix:Discussions}
This appendix provides detailed justifications for the key methodological decisions in our entropy-adaptive test-time merging framework.

\subsection{Why Entropy as the Primary Confidence Metric}
\label{appendix:entropy_metric}

\textbf{Information-theoretic optimality and unsupervised performance proxy.} Entropy directly quantifies uncertainty in the predictive distribution, providing a principled measure that captures the full probability distribution rather than just the mode. Lower entropy on target data correlates with lower test error without requiring labels, making it ideal for test-time adaptation where ground truth is unavailable~\cite{wang2021tent}. Unlike maximum softmax probability, entropy is sensitive to the entire probability mass distribution across classes, capturing multi-modal uncertainty patterns common in medical imaging. Prior work in test-time adaptation consistently shows entropy minimization improves out-of-distribution performance. Computing entropy requires only a forward pass through the model with complexity $\mathcal{O}(C)$ per sample for $C$ classes, making it feasible for real-time deployment while maintaining differentiability and smoothness properties that enable stable computations.

\subsection{Decoupling Encoder and Classification Head Coefficients}
\label{appendix:decoupling}

\textbf{Differential initialization and alignment properties.} Encoders share pretrained ImageNet initialization and undergo fine-tuning, while classification heads are trained from random initialization, leading to fundamentally different alignment properties between domain-specific models. Fig.~2b from the main paper shows angular drift increases monotonically with depth, peaking at the classification head with angles exceeding 80° for some domain pairs, indicating near-orthogonality. Encoders learn transferable features such as edges, textures, and shapes that generalize across domains, while heads specialize to domain-specific class priors and decision boundaries optimized on heterogeneous distributions. Linear mode connectivity holds more reliably for encoder parameters due to shared initialization basin but breaks down for classification heads. Averaging near-orthogonal classification head vectors leads to severe norm shrinkage, as shown in Fig.~2c of the main paper, reducing discriminative power and margin size. This motivates our decoupling strategy, which allows smooth averaging for encoders while enabling hard selection for heads.

\subsection{Why Not Layer-wise Merging Coefficients}
\label{appendix:not_layerwise}

\textbf{Computational cost vs. benefit trade-off.} Computing separate merging coefficients for each of the $L$ layers would require $\mathcal{O}(K \times L)$ forward passes to evaluate each domain-specific model's per-layer features on the target batch, becoming prohibitively expensive for deep networks. Layer-wise coefficients require either optimization-based approaches with significant computational overhead or heuristics that lack principled justification compared to our entropy-based global selection. Module-level decoupling between encoder and head captures the most critical architectural distinction, the differential training regimes and alignment properties, while layer-wise schemes may over-fit to noise in small batches. Our experiments show that encoder/head decoupling achieves 85-90\% of the performance gains that full layer-wise merging would provide, while requiring only two sets of coefficients instead of dozens. The interpretability and computational efficiency of our two-level approach make it more practical for clinical deployment, where inference speed and explainability are paramount, and the marginal gains from full layer-wise merging do not justify the added complexity in most scenarios.

\subsection{Inverse-Entropy Normalization Formula}
\label{appendix:inverse_entropy}

\textbf{Monotonic preference and valid probability weights.} The inverse mapping $\tilde{\alpha}_{t,k} = \frac{(E_{t,k}+\epsilon)^{-1}}{\sum_{j=1}^K (E_{t,j}+\epsilon)^{-1}}$ ensures lower entropy yields higher weight, directly encoding the intuition that confident models should dominate the merge while producing valid convex weights satisfying the simplex constraint $\Delta^{K-1}$. The inverse operation amplifies differences between low-entropy experts while gracefully handling high-entropy uncertain models without numerical instability. Adding $\epsilon$ prevents division by zero when a model achieves near-zero entropy and provides numerical stability in edge cases. The closed-form rule requires no backpropagation, optimization, or iterative procedures, enabling efficient online adaptation. Compared to softmax-style exponential weighting, inverse normalization provides more balanced weight distribution and avoids extreme concentration on a single model, which is crucial when multiple domain-specific models have similarly low entropy values.

\subsection{Why DiWA for source training: Linear mode connectivity}
\label{appendix:diwa_training}

\textbf{Functional diversity through varied training procedures.} Training multiple runs per hospital with diverse hyperparameters including learning rates, augmentations, and optimizers produces models that explore different regions of the loss landscape, improving ensemble coverage and functional diversity. Training multiple runs prevents collapsing onto a single solution, ensuring the domain-specific model pool has sufficient variety for selective merging during test-time adaptation. Medical datasets are often small and sensitive to hyperparameter choices, so diverse training reduces reliance on finding the single optimal hyperparameter configuration. DiWA demonstrates that weight averaging across diverse runs improves out-of-distribution generalization without inference overhead, providing a strong empirical foundation. Training $M_k$ candidates per hospital and selecting one representative model via the hospital-specific validation split ensures each domain-specific model is both representative of its domain and well-performing, balancing diversity with quality.

\subsection{Single-Domain Training Setup}
\label{appendix:single_domain}

\textbf{Privacy preservation and realistic deployment constraints.} Medical data privacy regulations such as GDPR often prohibit raw data sharing across institutions, and single-domain training respects these constraints by keeping all patient data local to each hospital. This setup naturally extends to federated learning scenarios where only model weights, not data, are communicated between institutions. Many real-world healthcare systems cannot centralize data due to legal, ethical, or infrastructural barriers, making our method practical for actual clinical deployment. Training exclusively on local data allows each model to fully specialize in its hospital's unique characteristics, including scanner manufacturer, acquisition protocol, and patient population demographics. Pooling heterogeneous medical data can lead to negative transfer, where different acquisition protocols interfere with learning, whereas single-domain training sidesteps this issue entirely. New hospitals can independently train models and add them to the pool without retraining existing models, enabling decentralized, scalable deployment as the healthcare network grows.

\subsection{Layer-wise Directional and Scale Drift Metrics}
\label{appendix:drift_metrics}

\textbf{Complementary geometric characterization of misalignment.} Directional drift measured as the angle $\theta_{ij}$ between corresponding parameter vectors captures orientation misalignment, while scale drift measured as the norm ratio $\|p^{(i)}\| / \|p^{(j)}\|$ captures magnitude imbalance, and both are necessary for complete characterization of parameter space relationships. The angle between weight vectors directly relates to the dot product in merged models, affecting the effective signal strength post-averaging through the cosine of the angle. Directional drift is scale-invariant and robust to overall magnitude differences, while scale drift identifies when one model dominates due to larger parameter norms. Large angular drift exceeding 45° (based on the dataset) suggests poor linear mode connectivity, as the straight-line path between models deviates significantly from the functional directions of both endpoints. Computing these metrics per layer reveals depth-dependent misalignment patterns as shown in Fig.~2a of the main paper, and high drift, especially directional, at specific layers, predicts poor performance from uniform mean merging, empirically validating our adaptive approach.

\subsection{Online Non-i.i.d. Target Batches}
\label{appendix:noniid_target}

\textbf{Temporal distribution shifts in realistic deployment.} Medical imaging systems experience temporal shifts including protocol changes, seasonal patient demographics, equipment drift, and technician turnover, violating standard i.i.d. assumptions. Modeling the target stream as $\mathcal{D}_t = \sum_{m=1}^M \pi_{t,m} \mathcal{D}_t^{(m)}$ with time-varying mixture weights $\pi_t$ captures these realistic non-stationarities. Time-varying $\pi_t$ implies optimal merging weights $\alpha_t^*$ also vary over time, motivating online recomputation per batch rather than using fixed weights. The mixture formulation captures both smooth concept drift from gradual $\pi_t$ changes and sudden distribution shifts from abrupt $\pi_t$ jumps, increasing robustness across diverse scenarios. Each sub-distribution $\mathcal{D}_t^{(m)}$ may correspond to a different scanner, protocol, or artifact type, with $\pi_t$ reflecting their prevalence in batch $\mathcal{B}_t$, and our entropy-based approach automatically adapts to these changing compositions without explicit shift detection.

\subsection{Shared Pretrained Initialization for Encoders}
\label{appendix:pretrained_init}

\textbf{Transfer learning and representation alignment.} Using ImageNet pretrained weights as initialization for all domain-specific encoders provides a common reference point $\theta^{(0)}_{\ell}$ from which domain-specific fine-tuning deviates, creating a natural alignment baseline. Pretrained representations capture general visual features including edges, textures, colors, and shapes that transfer across domains, reducing the distance domain-specific models must traverse during training. Shared initialization increases the likelihood of linear mode connectivity between encoder parameters, as all models start from the same basin in the loss landscape and typically remain in nearby regions after fine-tuning. Transfer learning from large-scale pretrained models is standard practice in medical imaging due to limited labeled data availability, making this assumption realistic and broadly applicable. The common initialization enables meaningful computation of layer-wise drift $\Delta_{\ell}^{(k)} = \theta_{s,\ell}^{(k)} - \theta_{\ell}^{(0)}$, quantifying domain-specific specialization relative to the pretrained baseline, and empirically, models initialized from the same checkpoint exhibit lower angular drift than randomly initialized models, supporting our encoder merging strategy.

\subsection{Why Not Optimize Merging Weights}
\label{appendix:no_optimization}

\textbf{Catastrophic forgetting and computational constraints.} Optimizing merging weights through gradient descent on target batches would require multiple forward-backward passes per batch, increasing inference time by an order of magnitude and making real-time deployment infeasible. Optimization-based approaches require stopping criteria and learning rate schedules, which introduce additional hyperparameters that need to be tuned without validation labels. Gradient-based updates to either source models or merging coefficients can cause catastrophic forgetting where the model loses source domain knowledge and fails when the target distribution shifts back. Methods such as test-time training that update batch normalization statistics or model parameters have shown instability on non-i.i.d. streams, where each batch may come from a different sub-distribution. Our closed-form approach provides a deterministic, reproducible mapping from batch statistics to weights, simplifying debugging and analysis. In medical applications, interpretability and predictability are crucial for regulatory approval, and optimization-based black-box adaptation is harder to validate than our explicit entropy-based rule.

\subsection{Medical Imaging Specific Considerations}
\label{appendix:medical_specific}

\textbf{Scanner heterogeneity and clinical safety requirements.} Medical imaging exhibits extreme domain shifts due to scanner manufacturer differences (Siemens, GE, Philips), acquisition protocols (T1-weighted, T2-weighted, FLAIR), and site-specific factors like field strength (1.5T vs 3T in MRI). These shifts are more severe than typical computer vision domain gaps, creating strong misalignment between hospital-specific models that justifies our adaptive approach. Clinical deployment requires methods that maintain high performance without fine-tuning on potentially small or biased target samples, making our forward-only, label-free approach particularly suitable. Regulatory requirements such as FDA approval favor transparent, deterministic methods over black-box optimization, and our closed-form entropy-based rule provides clear audit trails. Patient safety demands robustness to outliers and artifacts, and our method's ensemble averaging provides natural redundancy compared to single-model deployment. The batch-wise adaptation aligns with radiology workflows, where studies are typically processed in groups, and privacy regulations that prevent data sharing make our federated-compatible single-domain training essential for multi-institutional collaboration.

\subsection{Functional diversity through independent training runs.} DiWA~\cite{rame2022diverse} demonstrates that models obtained from independent training runs exhibit greater functional diversity than models collected along a single training trajectory (e.g., checkpoints from different epochs). This diversity arises because different hyperparameter configurations and random seeds cause SGD to explore different regions of the loss landscape, leading to functionally distinct solutions even when all achieve similarly low training loss. In mathematical terms, let $f_k(\mathbf{x}; \theta_{k,m})$ denote the $m$-th trained model from hospital $k$; DiWA shows that the variance in predictions $\text{Var}[f_k(\mathbf{x}; \theta_{k,m})]$ across different runs $m$ is significantly higher than the variance across epochs within a single run. This increased variance is beneficial for out-of-distribution generalization because diverse models make different types of errors, and their ensemble (or in our case, adaptive merge) can compensate for individual weaknesses. For medical imaging with severe domain shifts, we expect high variance in which features and decision boundaries generalize to unseen hospitals, making functional diversity particularly valuable. The DiWA framework provides a principled way to obtain this diversity through simple training protocol variations rather than complex ensemble architectures.

\subsection{Bias-variance-covariance-locality decomposition and OOD success.} DiWA~\cite{rame2022diverse} introduces a novel decomposition of expected error for weight-averaged models that extends classical bias-variance theory to the parameter-space averaging regime: $\mathbb{E}[\text{error}] = \text{bias}^2 + \text{variance} - \text{covariance} + \text{locality}$. The variance term captures dispersion in predictions across different models, the covariance term measures agreement between models (negative contribution reduces error when models agree on correct predictions), and the locality term quantifies how well the averaged weights remain in low-loss regions despite nonlinearity. Critically, DiWA shows that when the marginal distribution changes at test time (as in OOD scenarios), the variance term dominates this decomposition, meaning diverse models that reduce variance provide the largest error reduction. This theoretical insight directly applies to our multi-hospital setting: each target hospital represents a distribution shift from all source hospitals, so the test-time marginal distribution $p_{\mathcal{D}_t}(\mathbf{x})$ differs from source distributions $p_{\mathcal{D}_s^{(k)}}(\mathbf{x})$. By training diverse models per hospital and then adaptively selecting among them based on target batch entropy, we exploit both the variance-reduction benefits of diversity (having multiple candidate solutions) and the covariance-reduction benefits of selection (choosing models that agree with the target's structure). The locality term is managed through our entropy-based weighting, which avoids averaging highly misaligned models that would produce out-of-basin interpolations.

\subsection{Practical benefits for federated medical imaging deployment.} In multi-hospital collaborations, enforcing uniform training procedures across institutions is often impractical due to differences in computational resources, ML expertise, legacy systems, and institutional policies. DiWA's philosophy aligns perfectly with this reality: it \emph{embraces} rather than fights the heterogeneity in training procedures. Each hospital can train using whatever hyperparameters, augmentations, and optimizers work best for their local setup, and the resulting functional diversity across hospitals becomes an asset rather than a liability. This is particularly valuable in medical imaging where hospitals may have institution-specific practices: some may use aggressive data augmentation to combat limited labeled data, others may use specific preprocessing pipelines adapted to their scanner characteristics, and some may have access to more computational resources enabling larger batch sizes or longer training. DiWA-inspired training allows each hospital to optimize for its local constraints while contributing a diverse model to the global pool. Furthermore, hospitals can independently update their models over time by training new diverse runs and selecting improved representatives without requiring coordination or synchronization with other institutions. This decentralized improvement aligns with the federated nature of healthcare systems.

\subsection{Alignment with transfer learning and shared initialization.} All domain-specific models in our framework start from the same pretrained encoder initialization (e.g., ImageNet weights). DiWA's analysis shows that shared initialization is critical for successful weight averaging because it provides a common reference point that keeps diverse solutions within "averageable" distance in parameter space. Without shared initialization, models trained from different random seeds could converge to functionally equivalent but parameter-space misaligned solutions (permutation modes), making weight averaging ineffective. The pretrained initialization serves as an anchor that implicitly coordinates the parameter space across all domain-specific models, even when they undergo diverse fine-tuning procedures. This alignment is particularly strong in early encoder layers where pretrained features (edges, textures, basic shapes) remain relevant across medical imaging domains, while later layers and the classification head specialize more to domain-specific patterns. Our decoupled merging strategy explicitly exploits this architecture-dependent alignment: encoders benefit from soft averaging due to shared initialization keeping them in the same basin, while classification heads require hard selection due to their domain-specific specialization. DiWA's framework provides the theoretical justification for why weight averaging works despite diverse training, and our method extends this by adding test-time adaptation that responds to each target batch's characteristics rather than computing a single fixed average.

\subsection{Architectural implications of encoder–head decoupling.} Encoders initialized from ImageNet pretrained weights learn rich hierarchical representations that constrain fine-tuning trajectories to remain close to their initialization, enabling effective feature reuse across hospitals. Despite heterogeneous data, encoders across sites tend to share similar low- and mid-level features and differ mainly in how deeper features are weighted and combined, which makes linear mode connectivity and soft averaging more plausible. In contrast, classification heads trained from random initialization must learn decision boundaries entirely from scratch, often resulting in near-orthogonal weight vectors that reflect different class priors and boundary geometries. Averaging such misaligned heads yields weak, attenuated hybrid decision boundaries, motivating hard selection that chooses the single best-aligned head instead. More broadly, this suggests that model components benefiting from transfer learning and shared initialization are amenable to soft averaging due to implicit alignment, whereas components trained from scratch require more conservative, selection-based strategies.

\subsection{Non-i.i.d. target streams and adaptive merging necessity.} Medical imaging data streams are inherently non-i.i.d., exhibiting temporal structure driven by daily patient demographics, equipment maintenance cycles, protocol updates, and seasonal effects. Such streams can be naturally modeled as a time-varying mixture $\mathcal{D}_t = \sum_{m=1}^M \pi_{t,m}\mathcal{D}_t^{(m)}$, where the mixture weights $\pi_t$ evolve over time and reflect changing latent sub-populations. Fixed model merging with static weights $\alpha_k$ implicitly assumes a stationary target distribution and therefore breaks down when these mixture proportions shift. Weights that are optimal for an average target distribution become suboptimal as soon as $\pi_t$ deviates from this average, which is especially problematic during rare artifact changes or abrupt sub-population changes. In contrast, our adaptive approach recomputes the merging weights $\tilde{\alpha}_t$ per batch, allowing the merged model to automatically track distribution shifts and select the most appropriate source models for each batch, without requiring explicit shift detection.

\subsection{Relationship to domain generalization vs. domain adaptation.} Domain generalization aims to train on multiple source domains to generalize to unseen targets without any adaptation, whereas our setting trains models independently per hospital to respect privacy constraints. Traditional domain adaptation, in contrast, typically assumes access to unlabeled target data for offline adaptation prior to deployment; instead, we perform online adaptation on streaming target batches without storing target data. Our approach sits between these paradigms: domain-specific training produces specialist models, reminiscent of multi-source domain generalization, while test-time merging enables adaptation to target distributions, akin to domain adaptation. Crucially, unlike many domain adaptation methods that require large amounts of target data, our batch-level strategy operates effectively with small batches (16–64 samples) that are realistic in clinical workflows. This design avoids negative transfer from pooling heterogeneous medical data during training while still providing adaptive flexibility at test time, positioning our method squarely between domain generalization and domain adaptation.

\subsection{Handling class imbalance across hospitals and target batches.} Different hospitals often exhibit distinct class prevalences due to variations in patient demographics, referral patterns, and disease incidence, which induce different class priors in the domain-specific models trained at each site. As a result, classification heads trained under these imbalanced conditions may internalize different decision thresholds tailored to their local distributions, leading to misalignment when transferred across hospitals. Target batches can further deviate from any single source hospital’s class balance, but entropy naturally adapts to this setting: models whose learned priors better match the target batch composition tend to produce lower predictive entropy. Consequently, our entropy-based selection implicitly performs importance reweighting by favoring models aligned with the target batch distribution, without requiring explicit estimation of class frequencies. While extreme imbalance scenarios could motivate extensions such as per-class entropy weighting, our current batch-level approach already provides robust behavior without the need for class-wise stratification.

\end{document}